\documentclass[10pt,twocolumn]{IEEEtran}
\usepackage{float}
\usepackage{graphicx}
\usepackage{epsfig}
\usepackage{amsmath}
\usepackage{bbm}
\usepackage{amssymb}
\usepackage{setspace}
\usepackage{wrapfig}
\usepackage{times,color}
\usepackage{caption}
\usepackage{subcaption}
\usepackage{algorithm}
\usepackage{algpseudocode}
\usepackage{epstopdf}
\usepackage{cite,footnote,xspace,syntonly,bm}

\usepackage{amsfonts}

\allowdisplaybreaks[4]
\input{mysymbol.sty}

\algnewcommand{\Inputs}[1]{%
  \State \textbf{Inputs:}
  \Statex \hspace*{\algorithmicindent}\parbox[t]{.8\linewidth}{\raggedright #1}
}
\algnewcommand{\Initialize}[1]{%
  \State \textbf{Initialize:}
  \Statex \hspace*{\algorithmicindent}\parbox[t]{.8\linewidth}{\raggedright #1}
}

\def \cN {\mathcal{N}}

\def \cS {\mathcal{S}}
\def \cU {\mathcal{U}}

\def \bX {{\bf X}}

\def \bU {{\bf U}}

\def \bPsi {{\boldsymbol \Psi}}

\def \bu {{\bf u}}

\def \bx {{\bf x}}

\def \bu {{\bf u}}

\def \by {{\bf y}}

\def \bbeta {{\boldsymbol \beta}}

\def \bpsi {{\boldsymbol \psi}}

 \long\def\symbolfootnote[#1]#2{\begingroup
 	\def\thefootnote{\fnsymbol{footnote}}
 	\footnote[#1]{#2}\endgroup} \psfull

\begin{document}

\title{\huge Online Categorical Subspace Learning \\
for Sketching Big Data with Misses}

\author{{\it Yanning Shen, \textit{Student Member}, \textit{IEEE},  Morteza Mardani, \textit{Member}, \textit{IEEE}\\
 and Georgios~B.~Giannakis, \textit{Fellow, IEEE}}}

\markboth{IEEE TRANSACTIONS ON SIGNAL PROCESSING,  SEPTEMBER 23, 2016~(SUBMITTED)}{}
\maketitle  
\symbolfootnote[0]{$\dag$ Work in this paper
	was supported by NSF grants 1514056, and 1500713. Parts of the
	paper were presented in the {\it 49th Conference on Information Sciences and Systems}, Baltimore, Maryland, March 2015.}

\symbolfootnote[0]{$\ast$Y. Shen and G. B. Giannakis are with the Dept.
	of ECE and the Digital Technology Center, University of
	Minnesota, Minneapolis, MN 55455. Tel/fax:(612)625-4287/625-2002; Emails:
	\texttt{\{shenx513,georgios\}@umn.edu}. Morteza Mardani is also with the Stanford University, 875 Blake Wilbur Drive Stanford, CA 94305-5847; Email: \texttt{morteza@stanford.edu}}



\vspace{-8mm}
\begin{abstract}
	With the scale of data growing every day, reducing the dimensionality (a.k.a. sketching) of high-dimensional data has emerged as a task of paramount importance. Relevant issues to address in this context include the {\it sheer volume} of data that may consist of {\it categorical} samples, the  typically {\it streaming} format of acquisition, and the possibly {\it missing} entries. To cope with these challenges, the present paper develops a novel categorical subspace learning approach to unravel the latent structure for three prominent categorical (bilinear) models, namely, Probit, Tobit, and Logit. The deterministic Probit and Tobit models treat data as quantized values of an analog-valued process lying in a low-dimensional subspace, while the probabilistic Logit model relies on low dimensionality of the data log-likelihood ratios. Leveraging the low intrinsic dimensionality of the sought models, a rank regularized maximum-likelihood estimator is devised, which is then solved recursively via alternating majorization-minimization to sketch high-dimensional categorical data `on the fly.' The resultant procedure alternates between sketching the new incomplete datum and refining the latent subspace, leading to lightweight first-order algorithms with highly parallelizable tasks per iteration. As an extra degree of freedom, the quantization thresholds are also learned jointly along with the subspace to enhance the predictive power of the sought models. Performance of the subspace iterates is analyzed for both {\it infinite} and {\it finite} data streams, where for the former asymptotic convergence to the stationary point set of the {\it batch} estimator is established, while for the latter {\it sublinear} regret bounds are derived for the empirical cost. Simulated tests with both synthetic and real-world datasets corroborate the merits of the novel schemes for {\it real-time} movie recommendation and chess-game classification.		
\end{abstract}

\begin{keywords}
	Categorical data, sketching, online subspace learning, rank regularization, regret analysis.
\end{keywords}



\section{Introduction}
Principal component analysis (PCA) is arguably the most popular tool for dimensionality reduction, with numerous applications in science and engineering \cite{Jolliffe2002}. It is however primarily designed to sketch high-dimensional data with analog-amplitude values, and does not suit categorical data emerging for instance, with recommender systems. Categorical PCA seeks a low-dimensional sketch of the high-dimensional categorical data to render affordable downstream machine learning tasks such as imputation, classification, and clustering; see e.g.,~\cite{Tipping98,ALL03,LHH10,De2006,kozma2009binary,JMP12,ZC14}. However, the growing scale of nowadays `Big Data' applications, such as  recommender systems (e.g., NetFlix) with millions of users rating thousands of movies, pose extra challenges: (c1) the sheer volume of data approaches the computational and storage limits; (c2) new releases demand real-time processing for recommendations; and (c3) absent data entries, corresponding to missing user ratings.

Past works on categorical PCA focus on binary PCA, and rely on logistic-regression entailing (bi)linear models; see e.g.,~\cite{ALL03,kozma2009binary,LHH10}. The work in \cite{ALL03} assumes that the log-odds matrix of the data lies in a linear low-dimensional subspace. The approach in \cite{kozma2009binary} further imposes a Gaussian prior on the sketch, whereas the one in \cite{LHH10} promotes sparsity for the subspace to regularize the log-likelihood, which is then maximized using a batch majorization-minimization (MM) scheme. In a similar vein, binary matrix factorization and binary dictionary learning have also been employed for dimensionality reduction when batch processing is affordable; see e.g., \cite{koyuturk2003proximus,shen2009mining,koyuturk2005compression, lu2011weighted,davenport20141,haupt2014sparse}. Other techniques for summarizing discrete-valued data include multidimensional scaling \cite{rohde2002methods}, and the $k$-modes algorithm \cite{huang1997fast}, which extends $k$-means~\cite{macqueen1967some} to the discrete domain by adopting a proper dissimilarity measure. For streaming datasets, \cite{ZC14} proposes an online sketching scheme based on logistic PCA when all data entries are present, and an online binary dictionary learning algorithm has been developed in \cite{shen16onlineDL}.  All in all, the prior art is for the most developed for binary data, and either assumes the data have no missing entries, or, it relies on batch processing.



To cope with challenges (c1)-(c3), the present paper brings forth a novel categorical subspace learning (CSL) scheme that unravels the latent structure behind the categorical data for three popular bilinear schemes; namely, Probit, Tobit, and Logit \cite{bishop06book}. The Probit model treats categorical data as quantized values of a certain analog-amplitude vector that lies in a linear low-dimensional subspace. Tobit is the model of choice for censoring, while the probabilistic Logit model generalizes logistic regression to the unsupervised case. The bilinear models in this paper can accommodate finite-alphabet datasets, and can also interpolate missing entries via rank regularization. To this end, the log-likelihood is regularized with a term corresponding to the rank of the underlying analog-valued data matrix. Leveraging a decomposable variant of the nuclear-norm, a recursive nonconvex program is then formulated, and solved online via stochastic alternating minimization. The resultant procedure alternates between sketching the new datum and refining the latent subspace via stochastic gradient descent to extract the information present in the new datum. This leads to lightweight first-order iterates that are nicely parallelizable across the latent subspace dimension, and thus implemented very efficiently via graphical processing units (GPUs).

The deterministic Probit and Tobit models adopt pre-determined quantization thresholds, which we further adjust to enhance the predictive power of the categorical models. To this end, the first-order iterates are modified to jointly learn the quantizer thresholds as well as the latent subspace. Performance of the subspace iterates is also analyzed for both {\it finite} and {\it infinite} data streams, where the former relies on martingale sequences to prove asymptotic convergence of the subspace to the stationary point set of the batch maximum likelihood (ML) estimator. For finite data streams, an unsupervised notion of regret is adopted to derive sublinear regret bounds for the empirical cost. Extensive simulated tests are performed with synthetic and real datasets for classification of chess-game scenaria, and interpolation of absent ratings in movie recommender systems. They corroborate the convergence and effectiveness of the novel sketching scheme in terms of accuracy and runtime relative to the existing alternatives.


The rest of this paper is organized as follows. Section~\ref{sec:pre} presents preliminaries, and states the problem. Section~\ref{sec:rank_reg} formulates the ML estimator with rank regularization, based on which Section~\ref{sec:online} develops subspace learning algorithms for online sketching via stochastic alternating minimization. Learning the quantizer is the subject of Section~\ref{sec:learn_quan}, while the performance of first-order subspace iterates is analyzed in Section~\ref{sec:perf_analysis}. Section \ref{sec:test} reports the numerical tests with synthetic and real datasets, while conclusions are drawn in Section \ref{sec:conc}.

\noindent\textit{Notation}: Bold uppercase (lowercase) letters will denote matrices (column vectors), and calligraphic letters will be used for sets, while operators $(.)^\top$, $\mathbb{E}(.)$, $\sigma_{\max}$ and $\sigma_{\min}$ will denote transportation, expectation, maximum, and minimum singular value, respectively. The $\ell_p$-norm of $\bbx \in \mathbb{R}^n$ is $\|\bbx\|_p:= (\sum_{i=1}^n |x_i|^p)^{1/p} $ for $p \geq 1$. For matrices $\bbA$, $\bbB\in \mathbb{R}^{m\times n}$, $\langle\bbA, \bbB\rangle:={\rm tr}(\bbA\bbB^{\top})$ is their trace inner product. and $\|\bbA\|_F:=\sqrt{{\rm tr}(\bbA\bbA^\top)}$ is the Frobenius norm, while $\|\bbA\|$ represents the spectral norm which corresponds to the largest singular value of the matrix; and, $\mathcal{I}(\epsilon)$ is the indicator function taking value $1$ if event $\epsilon$ holds, and $0$, otherwise.

\section{Preliminaries and Problem Statement}\label{sec:pre}
Consider the high-dimensional $D\times 1$ vectors $\{\bby_{\tau}\}_{\tau=1}^T$ with categorical entries drawn from a $J$-element alphabet $\mathcal{S}:=\{s_0,\ldots,s_{J-1}\}$. For instance, in movie recommender systems $\by_t$ represents the users' categorical ratings (e.g., ``good'' or ``bad'') for the $t$-th movie. Apparently, each user can only rate a small fraction of movies, and thus ratings for a sizable portion of movies may not be available. Let $\Omega_{t}\subseteq \{1,\ldots, D\}$ with cardinality $|\Omega_t|~(\ll D)$ denote the set of available entries (user ratings) associated with the $t$-th movie. With the {\it partial} categorical data $\{y_{t,i},~i \in \Omega_t\}_{t=1}^T\in \cS^D$, categorical PCA seeks a low-dimensional (sketched) set of features $\{\bbpsi_{\tau}\}_{\tau=1}^T\in\mathbb{R}^d$ (with $d \ll D$), which render 
affordable downstream inference tasks such as regression, prediction, interpolation, classification, or, clustering; see e.g.,~\cite{ALL03,LHH10,De2006,kozma2009binary,JMP12,ZC14}. Aiming at a related objective, the present work builds on three {\it unsupervised} categorical models that are described next.

\begin{figure*}[t]
	\centering
		\includegraphics[width=16cm]{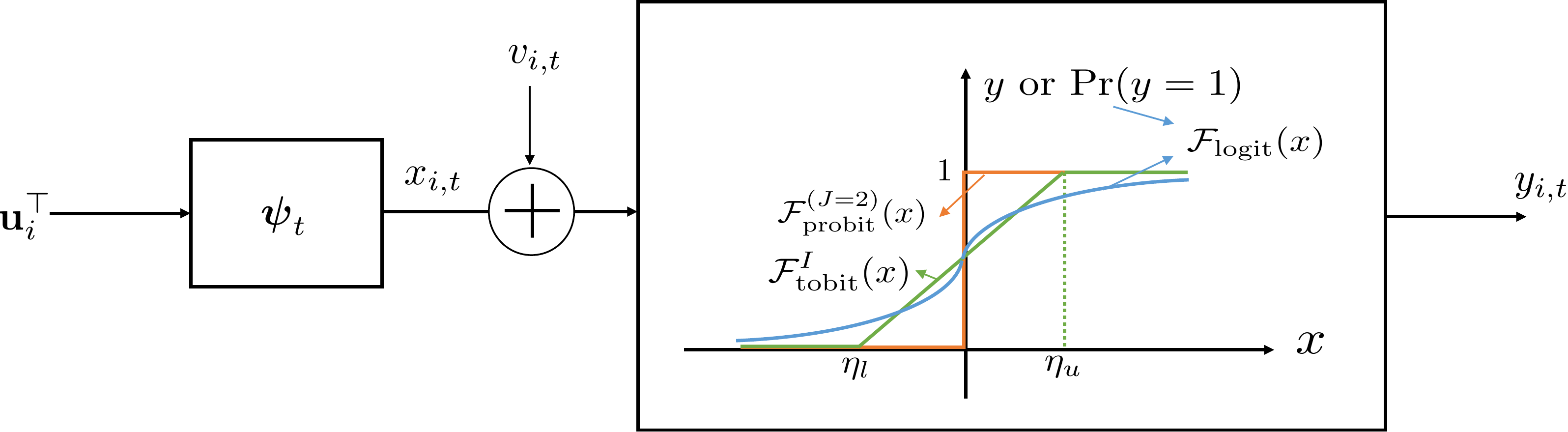}
	\vspace{1mm}
	\caption{Illustration of the considered models, namely Probit, Tobit and Logit.}\label{fig:model}
\end{figure*}

\subsection{Blind Probit model}\label{subsec:probit}
The Probit model regards $\cS$ as the range space of a $J$-element quantization mapping 
\begin{align}
\label{mod:probit}
\mathcal{F}^{(J)}_{\rm probit}(x):=s_j&~\text{ if }~x \in (\eta_j,\eta_{j+1}]\nonumber\\
&\hspace{1cm}\text{  for } j=0,1,\ldots, J-1
\end{align}
with  $\{\eta_j\}$ denoting known quantization thresholds. The categorical vectors $\{\bby_{t}\}_{t=1}^T$ are then viewed as the quantized versions of certain analog-valued data vectors that belong (or lie close) to a linear low-dimensional subspace $\cU$. Specifically, the $i$-th entry admits the following quantized bilinear model 
\begin{subequations}
\label{eq:model:probit}
\begin{eqnarray}
y_{i,t}&=&\mathcal{F}_{\rm probit}^{(J)}(x_{i,t}+v_{i,t})\\
x_{i,t} &:=& \mathbf{u}_i^{\top}\bbpsi_{t},\qquad i \in \Omega_{t}
\end{eqnarray}
\end{subequations}
where $\bbpsi_{t}\in\mathbb{R}^{d}$ denotes the projection of $\bby_t\in\mathbb{R}^D$ onto the low-dimensional ($d<D$) subspace $\cU$; see also Fig. \ref{fig:model}. Columns of the matrix $\bbU:=[\bu_1,\ldots,\bu_{D}]^{\top}$, where $\bbu_i^{\top} \in \mathbb{R}^{d}$ denotes the $i$-th row of $\bU$ span the linear subspace $\cU$. The noise $v_{i,t}$ also accounts for errors and unmodeled dynamics. 

Our goal of finding $\{\bbpsi_t\}_{t=1}^T$ and $\bbU$ corresponds to blind regression given finite-alphabet $\{y_{i,t}\}$, while for $\bbU$ known, it is closely related to nonblind Probit-based classification.


\subsection{Blind Tobit model}\label{subsec:tobit}
Acquired data in practice can be censored to e.g., lie in a prescribed range, for further processing. Given thresholds $\eta_l$ and $\eta_u$, a typical censoring rule discards large data entries based on
\begin{align}
\label{tobit1}
\mathcal{F}_{\rm tobit}^{I}(x):=\left\{
\begin{array}{ccl}
\eta_u     &      & {x\geq\eta_u}\\
\eta_l     &      & {x\leq\eta_l}\\
x       &      & {x\in(\eta_l,\eta_u)}.
\end{array} \right.
\end{align}
Alternatively, one can think of a censoring rule that removes small data entries as effected by
\begin{align}
\label{tobit2}
\mathcal{F}_{\rm tobit}^{\rm II}(x):=\left\{
\begin{array}{ccl}
x     &      & {x\geq\eta_u}\\
x     &      & {x\leq\eta_l}\\
\eta       &      & {x\in(\eta_l,\eta_u)}.
\end{array} \right.
\end{align}
%

To gain insight on the nonlinear maps \eqref{tobit1} and \eqref{tobit2}, consider the study of survival rates for patients with a certain disease during a certain time period. If the patient dies naturally within the study period, one knows precisely the survival time. However, if the patient dies before or after the study, where no accurate data is collected, only an upper bound or a lower bound is available on the patient age. Tobit models have been shown useful in big data applications for selecting informative observations~\cite{berberidis2015online}. 

Similar to \eqref{eq:model:probit}, one can postulate the censored bilinear Tobit model (see also Fig. \ref{fig:model})
\begin{subequations}
\label{eq:model:tobit}
\begin{eqnarray}
y_{i,t}&=& \mathcal{F}_{\rm tobit}(x_{i,t}+v_{i,t})\\
 x_{i,t}&:=&\bbu_i^\top\bbpsi_t ,\qquad i \in \Omega_{t}.
\end{eqnarray}
\end{subequations}

\subsection{Blind Logit model}\label{subsec:logit}
Probit and Tobit adopt deterministic data-generating functions $\mathcal{F}$ and rely on nonlinear regression to predict missing categorical (hard) data. Inspired by logistic regression, Logit relies on a {\it probabilistic} (soft) model to predict label probabilities~\cite{trevor2009elements}. Suppose $\{y_{i,t}\}$ are mutually independent random variables, where the $i$-th entry $y_{i,t}$ is Bernoulli distributed with success probability $\pi_{i,t}:={\rm Pr}(y_{i,t}=1)$. Define also the log-likelihood ratio $x_{i,t} := \log\{\pi_{i,t} /(1- \pi_{i,t} )\}$, which upon solving for $\pi_{i,t}$ yields the Logit function $\pi(x) := \{1 + \exp(−x)\}^{-1}$. 

The Logit model postulates that the log-likelihood ratio sequence $\{x_{i,t}\}$ belongs to a linear low-dimensional subspace spanned by the matrix $\bU$; that is, $x_{i,t}:=\bbu_i^\top\bbpsi_t$ for some $\bbpsi_t$, and for the binary case ($s\in\{0,1\}$), the categorical data probability is thus expressed as
\begin{eqnarray}
{\cal F}_{\rm logit}(x_{i,t})
&:=&{\rm Pr}(y_{i,t}=s)\nonumber\\
 &=&\frac{1}{1+\exp((1-2s)x_{i,t})},~~ i \in \Omega_{t}.
\label{eq:model:logit}
\end{eqnarray}
Likewise for the multibit Logit with each entry chosen from a $J$-element alphabet, $J-1$ bilinear Logit models start from the log-likelihood ratio
\begin{align}
	\log \frac{{\rm Pr}(y_{i,t}=s_j)}{{\rm Pr}(y_{i,t}=s_0)}=\bbpsi_t^\top{\bbu_i^{(j)}}, \quad j=1,\ldots,J-1
	\label{eq:model:log}
\end{align}
where $\bbu_i^{(j)}$ is the predictor for the $j$-th class, and adopt the soft data model to arrive at (cf. \eqref{eq:model:logit})
\begin{align}
\label{eq:model:logitm}
	{\rm Pr}(y_{i,t}=s_j)=\frac{\exp({\bbpsi_t^\top{\bbu_i^{(j)}} } )}{1+\sum_{k=1}^{J-1}\exp({\bbpsi_t^\top}{\bbu_i^{(k)}})}, ~~ j=1,\ldots,J-1.
\end{align}
Different from \eqref{eq:model:probit} and \eqref{eq:model:tobit} where (hard) categorical data $y_{i,t}$ are nonlinear functions of $x_{i,t}$, Logit deals with (soft) probability data ${\rm Pr}(y_{i,t}=s)$, expressed in \eqref{eq:model:logitm} as a nonlinear function of $x_{i,t}$. 

Given $\{y_{i,t}\}$, the ensuing section will develop ML estimators of $\bbU$ and $\{\bbpsi_t\}$ for the three models introduced in this section, namely \eqref{eq:model:probit}, \eqref{eq:model:tobit}, and \eqref{eq:model:logitm}.

\section{Rank-Regularized ML Estimation}\label{sec:rank_reg}
In what follows the likelihood function will be derived first, when the additive noise $v_{i,t}\sim \mathcal{N}(0,\sigma^2)$ is independent and identically distributed (i.i.d.), zero-mean Gaussian, with variance $\sigma^2$. As a result, available categorical entries $\{y_{i,t}\}$ are independent across $i$ and $t$.

\subsection{Log-likelihood function}
For the Probit model in \eqref{eq:model:probit}, the per-categorical-entry likelihood can be written as  
\begin{align}
&{\rm Pr} (y_{i,t};\bbu_i,\bbpsi_t)=\prod_{j=0}^{J-1}{\rm Pr}\{x_{i,t}\in(\eta_{j},\eta_{j+1}]\}^{\mathcal{I}(y_{i,t}=s_j)}\nonumber\\
=&\prod_{j=0}^{J-1} \left[{\rm Q}\left( \frac{\eta_{j}-\mathbf{u}_i^{\top}\bbpsi_{t}}{\sigma} \right)-{\rm Q} \left(\frac{\eta_{j+1}-\mathbf{u}_i^{\top}\bbpsi_{t}}{\sigma} \right)\right]^{\mathcal{I}(y_{i,t}=s_j)}
\label{eq:prob_y}
\end{align}
where $\mathcal{I}(\epsilon)$ is the indicator function, and ${\rm Q}(\cdot)$ denotes the standard Gaussian tail function. Upon collecting the low-dimensional representations in a matrix $\bbPsi :=[\bbpsi_1,\ldots,\bbpsi_{T}]$, the log-likelihood of the available categorical data can be expressed as
\begin{subequations}
\label{eq:log_lik}
\begin{eqnarray}
	&&\hspace{-2cm}\log \mathcal{L}_{\rm probit}\Big(\{y_{i,\tau},~i \in \Omega_{\tau}\}_{\tau=1}^T;\bbU,\bbPsi \Big) \nonumber 
	\\&=&\sum_{\tau=1}^T\sum_{i\in\Omega_{\tau}}\log\ell_{\rm probit}(y_{i,\tau}; \bbu_i,\bbpsi_{\tau})  \label{eq:log_lik_prob}
\end{eqnarray}
with
\begin{align}	
	&\log\ell_{\rm probit}(y_{i,t};\bbu_i,\bbpsi_{t})
:=\sum_{j=0}^{J-1} \mathcal{I}(y_{i,t}=s_j)\nonumber\\
\times &\log   \left[ {\rm Q}\left( \frac{\eta_{j}-\mathbf{u}_i^{\top}\bbpsi_{t}}{\sigma} \right)-{\rm Q}\left(\frac{\eta_{j+1}-\mathbf{u}_i^{\top}\bbpsi_{t}}{\sigma} \right)\right] \;.
\end{align}
	\end{subequations}

%
For the Tobit-I model in \eqref{tobit1}, one can readily derive the per-entry log-likelihood as 
\begin{subequations}
\label{l:tobit}
\begin{align}
\ell_{\rm tobit-I}(y_{i,t};\bbu_i,\bbpsi_{t})&:=  {\rm \phi}\left(\frac{y_{i,t}-\bbu_i^\top\bbpsi_t}{\sigma}\right) \mathcal{I}( y_{i,t}\in(\eta_l,\eta_u)) \nonumber\\ 
+& {\rm Q}\left(\frac{\eta_u-\bbu_i^\top\bbpsi_t}{\sigma}\right) \mathcal{I}(y_{i,t}=\eta_u) \nonumber\\ 
+& \Big[1-{\rm Q}\left(\frac{\eta_l-\bbu_i^\top\bbpsi_t}{\sigma}\right)\Big]  \mathcal{I}(  y_{i,t}=\eta_l) \label{eq:l:tobit}
\end{align}
with ${\rm \phi}(\cdot)$ denoting the probability density function (pdf) of the standardized Gaussian $\mathcal{N}(0,1)$. 

Likewise, the corresponding log-likelihood for the Tobit-II in \eqref{tobit2} can be represented as
\begin{align}
&\ell_{\rm tobit-II}(y_{i,t};\bbu_i,\bbpsi_{t})\nonumber\\
:= & \left[ {\rm Q}\left( \frac{\eta_{j}-\mathbf{u}_i^{\top}\bbpsi_{t}}{\sigma} \right)-{\rm Q}\left(\frac{\eta_{j+1}-\mathbf{u}_i^{\top}\bbpsi_{t}}{\sigma} \right)\right]\mathcal{I}( y_{i,t}=\eta) \nonumber\\ 
&+{\rm \phi}\left(\frac{y_{i,t}-\bbu_i^\top\bbpsi_t}{\sigma}\right)  \mathcal{I}(y_{i,t}\geq\eta_u) \nonumber\\
&+{\rm \phi}\left(\frac{y_{i,t}-\bbu_i^\top\bbpsi_t}{\sigma}\right)  \mathcal{I}(y_{i,t}\leq\eta_l).  \label{eq:l:tobit2}
\end{align}
\end{subequations}
 The overall log-likelihood for censored data is then obtained similar to~\eqref{eq:log_lik_prob}.

Finally, for the Logit model, based on the per-datum likelihood in \eqref{eq:model:logitm}, the per-entry log likelihood can be written as
\begin{align}
\label{l:logit}
&\ell_{\rm logit}(y_{i,t};\bbu_i,\bbpsi_{t})\nonumber\\
&\hspace{-0.2 cm}:=\sum_{j=0}^{J-1}  \mathcal{I}(y_{i,t}=s_j) \log   \left[ \frac{\exp({\bbpsi_t^\top{\bbu_i^{(j)}} } )}{1+\sum_{k=0}^{J-1} \exp({\bbpsi_t^\top{\bbu_i^{(k)}} } )}\right]
\end{align}
and consequently the overall log-likelihood can be obtained by substituting \eqref{l:logit} into the counterpart of \eqref{eq:log_lik_prob}, where Probit is replaced by Logit. 

So far \eqref{eq:log_lik}, \eqref{l:tobit}, and \eqref{l:logit} provide the building blocks of our ML criterion for the Probit, Tobit, and Logit model, respectively. In our ML approach however, we have not yet accounted for the low-rank property inherent to our data $\{y_{i,t}\}$, or, their probabilities $\{{\rm Pr}(y_{i,t}=s_j)\}$. This is the subject dealt with in the next subsection.



\subsection{Rank-regularized criterion}
%
Collect entries $x_{i,t}=\bbu_i^\top\bbpsi_t$ to form the $D\times 1$ vector $\bx_t=\bU \bpsi_t$. Since the stream $\{\bx_t\}$ lies in a linear low-dimensional subspace, $\bX:=[\bx_1,\ldots,\bx_T]=\bbU\bPsi$ is a low-rank matrix. A natural way to account for this property is to constrain the likelihood maximization over the set of low-rank matrices. However, since minimizing rank is in general NP-hard, the nuclear norm $\|\bbX\|_{*}:=\sum_i\sigma_i(\bbX)$ (where $\sigma_i$ signifies the $i$-th singular value) will be adopted as a convex surrogate for the rank~\cite{fazel}. These considerations prompted us to minimize the regularized negative log-likelihood
\begin{align}
\text{(P1) }~~~\min _{\bbX=\bbU\bbPsi}-\log\mathcal{L}\Big(\{y_{i,\tau},~i \in \Omega_{\tau}\}_{\tau=1}^T;\bbU,\bbPsi\Big) \hspace{-1mm}+\hspace{-1mm}\lambda \|\mathbf{X}\|_{*}\nonumber
\end{align}
where $\mathcal{L}$ collectively refers to the likelihood for any of the models in \eqref{eq:model:probit}, \eqref{eq:model:tobit}, or \eqref{eq:model:log}. The parameter $\lambda$ also controls the dimension of the latent subspace, and it can be tuned using cross validation. For the binary case $(J=2)$, the nuclear-norm regularization in (P1) has been shown under mild conditions to offer reconstruction guarantees for the Probit and Logit models~\cite{davenport20141}.

Apparently, the regularizer in (P1) entangles the data points, and as a result it challenges the development of efficient online solvers. To mitigate this computational challenge, the following bilinear characterization of the nuclear-norm is adopted (cf. \cite{MMG13JSTSP,mardani2015subspace,shen15})    
\begin{align} 
&\|\bbX\|_{*} =  \min_{\{\bbU,\bbPsi\}} \frac{1}{2} \left( \|\bbU\|_F^2 + \|\bbPsi\|_F^2 \right) \nonumber\\
&\text{s.~to}\quad \bbX= \bbU \bbPsi \label{eq:nuc}
\end{align}
where the minimization is over all possible bilinear factorizations of $\bbX$. Bypassing the need for calculating singular values of $\bbX$ whose size grows with time, this characterization of the nuclear norm not only effects a surrogate of the rank constraint, but also decouples variables across time, thus facilitating online optimization tasks~\cite{MMG13JSTSP,mardani2015subspace}. Utilizing \eqref{eq:nuc} into (P1) after dropping the $\min$ operation, yields
\begin{align}
&\text{(P2) }~~~~~\quad\min _{\{\bbU,\bbPsi\}}-\log\mathcal{L}\Big(\{y_{i,\tau},~i \in \Omega_{\tau}\}_{\tau=1}^T;\bbU,\bbPsi\Big) \nonumber \\ 
&\hspace{4.5cm} +\frac{\lambda}{2} \left( \|\bbU\|_F^2 + \|\bbPsi\|_F^2 \right).\nonumber
\end{align}
%


Since the $\min$ operation is in effect at the optimum, it can be easily seen that the solutions of (P2) and (P1) coincide~\cite{MMG13JSTSP}. For a moderate number of data entries $D$ and instants $T$, if
the entire data is available in batch, one can develop alternating
minimization algorithms along the lines of~\cite{MMG13JSTSP}. This amounts to cycling over two groups of variables, namely $\{\bU,\bPsi\}$, to jointly refine the sketch $\bPsi$ and the subspace $\bU$. However, for `Big Data' applications with ($D\gg$) streaming over time ($T \rightarrow \infty$), the size of $\bPsi$ grows;
thus, batch solvers become prohibitively complex, which well motivates  the recursive solvers of the ensuing section.


\section{Online Categorical Subspace Learning}\label{sec:online}
With modern `Big Data' applications, the massive amount of available data makes it impractical to store and process the data in an offline fashion. Furthermore, in many settings, the data are acquired sequentially over time and there is a need for real-time processing. In either case, practical limitations call for online schemes, capable of refining the sketch by adjusting the learned subspace to each new datum `on the fly.' With this in mind, we recast (P2) to minimize the following empirical cost
\begin{align}
\text{(P3)}\qquad\min _{\{\bbpsi_{\tau}\}_{\tau=1}^t, \bbU }\quad\frac{1}{t} \sum_{\tau=1}^t g_{\tau} \Big(\{y_{i,\tau}\}_{i\in\Omega_{\tau}};\bbpsi_{\tau},\mathbf{U}\Big) \nonumber
\end{align}
where the instantaneous cost $g_{\tau}$ corresponding to the $\tau$-th datum is given by
{\small
\begin{align}
&g_{\tau}\big(\{y_{i,\tau}\}_{i\in\Omega_{\tau}};\bbpsi_{\tau},\mathbf{U}\big) \nonumber \\ 
\hspace{-3mm} &:=-\sum_{i\in\Omega_{\tau}}\log\ell(y_{i,\tau};\bbpsi_{\tau},\mathbf{u}_i)+\frac{\lambda}{2t}\sum_{i=1}^D\|\mathbf{u}_i\|_2^2+\frac{\lambda}{2}\|\bbpsi_{\tau}\|_2^2. \label{eq:inst_loss}
\end{align}
}
It is important to recognize that different from our schemes in~\cite{MMG13JSTSP} and~\cite{mardani2015subspace}, which rely on analog-valued data, the nonlinear cost in (P3) entails categorical data and Gaussian tail functions that challenge algorithmic derivations. This is further elaborated next. 


\subsection{First-order alternating minimization algorithms}
To effectively solve (P3) for streaming data, an iterative alternating minimization (AM) method is adopted, where the iteration index coincides with the acquisition time. The sought AM scheme comprises two learning steps. Upon acquiring $\{y_{i,t}\}_{i \in \Omega_t}$ at time instant $t$, the first step (S1) embeds the data into the latent low-dimensional subspace, updates the features $\bbpsi_t$, and as a byproduct imputes the missing data entries. Subsequently, step (S2) refines the latent subspace according to the latest imputed datum. 

In (S1), given the subspace at the previous update $\bU[t-1]$, the embedding is obtained as
\begin{align}
\bbpsi_t= \arg\min_{\bbpsi\in\mathbb{R}^d} g_{t}\big(\{y_{i,t}\}_{i \in \Omega_t};\bbpsi,\mathbf{U}[t-1]\big). \label{eq:online1}
\end{align}

This amounts to a nonlinear ridge-regression task, given categorical $\{y_{i,t}\}_{i \in \Omega_t}$ with misses, along with their predictors $\{\bu_i[t-1]\}_{i\in\Omega_t}$ corresponding to the rows of $\bU[t-1]$. In the binary Probit model, the embedding $\bbpsi_t$ can also be viewed as the classifying hyperplane that assigns vectors $\bu_i[t-1],~i\in\Omega_t$, to their labels. With this interpretation, the $j$-th absent entry can be imputed by projecting $\bu_j[t-1]$ onto the hyperplane $\bbpsi_t$ that is then quantized to return the label $\text{sign}(\bu_j^{\top}[t-1] \bbpsi_t)$. Similarly, if the Logit model is adopted, \eqref{eq:online1} can be viewed as training a binary logistic regression classifier. 

The optimization problem \eqref{eq:online1} involves only $d \ll D$ variables, and can be readily solved using off-the-shelf solvers, such as gradient descent or Newton method. The recursions for the Probit model  are derived after regularizing \eqref{eq:log_lik} as in \eqref{eq:inst_loss}, and the corresponding iterates are listed in Algorithm \ref{algo:online}.

\begin{algorithm}[t] 
	\caption{Online rank-regularized ML sketching for the Probit model}\label{algo:online}
	\begin{algorithmic} 
		\State\textbf{input:}~$\{y_{i,\tau},~i \in \Omega_{\tau}\}_{\tau=1}^T$, $\{\mu_{t}\}$, $\lambda$
		
		\State\textbf{initialize}~$\bU[0]$ at random.
		
		\For {$t = 1, 2,\ldots$}
		\vspace{1.5mm}
		
		\State  {\bf(S1)} Sketching via first-order Algorithm~\ref{alg:psi:probit} or 
		\State second-order Algorithm \ref{alg:psi:probit2}
		
		\vspace{1.5mm}
		
		\State $\boldsymbol{\psi}_t= \arg\min_{\boldsymbol{\psi}\in\mathbb{R}^d} g_{t}\big(\{y_{i,t}\}_{i \in \Omega_t};\bbpsi,\mathbf{U}[t-1]\big)$
		
		\vspace{2mm}
		
		\State {\bf(S2)} Parallel subspace refinement 
		\vspace{2mm}
		\State $z_{j,t-1}^i:={\sigma}^{-1}(\eta_{j}-\mathbf{u}_i^{\top}[t-1]\bbpsi_t)$
		\vspace{2mm}
		
		\State $f_{i,t}:=\sum_{j=0}^{J-1}\mathcal{I}(y_{i,t}=s_j)\left[{\rm \phi}(z_{j,t-1}^i)-{\rm \phi}(z_{j+1,t-1}^i)\right]$

		\State 	$w_{i,t}:=\sum_{j=0}^{J-1}\mathcal{I}(y_{i,t}=s_j)\left[{\rm Q}(z_{j,t-1}^i)-{\rm Q}(z_{j+1,t-1}^i)\right]$
		
		
		
		\vspace{1mm}
		
		\State 
		$
		\mathbf{u}_i[t]=
		\left\{
		\begin{tabular}{ll}
	 	\hspace{-3mm}$(1-\lambda \mu_{t}/t)\mathbf{u}_i[t-1] + \mu_{t}({f_{i,t}}/{w_{i,t}}) \bbpsi_t$, & \hspace{-3.5mm}$i\in\Omega_t$\\
	\hspace{-3.mm}$(1-\lambda \mu_{t}/t)\mathbf{u}_i[t-1]$, &\hspace{-3.5mm}$i\notin \Omega_t$\\
  \end{tabular}
  \right.
  $

		\vspace{1mm}
		
		\EndFor\\
		\Return $\Big(\bU[t],\{\bbpsi_{\tau}\}_{\tau=1}^t \Big)$
		
	\end{algorithmic}
\end{algorithm}

\setcounter{algorithm}{0}
\newcounter{subsec}
\setcounter{subsec}{1}
\renewcommand\thealgorithm{\arabic{algorithm}\alph{subsec}}
\begin{algorithm}[t]
	\caption{Gradient-descent algorithm to obtain the sketch for the Probit model} \label{alg:psi:probit}
	\begin{center}
		\begin{algorithmic} 
			\State\textbf{input:}~$\{y_{i,t},~i \in \Omega_{t}\}$, $\{\eta_j\}$, $\{\beta_k\}$, $\lambda$, $K$, $\sigma$, $\bbU[t-1]$
			
			\State\textbf{initialize:}~$\bbpsi_t^{(0)}$
			\For{$k = 1, \ldots,K$}

			\vspace{1.5mm}
			
			
			\vspace{1.5mm}
			\State $a_{j,t-1}^i={\sigma}^{-1}(\eta_{j}-\mathbf{u}_i^{\top}[t-1]\bbpsi_t^{(k-1)})$
			
			\State $\epsilon_{i}:=\sum_{j=0}^{J-1}\mathcal{I}(y_{i,t}=s_j)\left[\frac{{\rm \phi}(a_{j,t}^i)-{\rm \phi}(a_{j+1,t}^i)}{{\rm Q}\left(a_{j,t-1}^i\right)-{\rm Q}(a^i_{j+1,t-1})}\right]$
			\vspace{1.5mm}
			\State 
			$\bbpsi_t^{(k)}=(1-\beta_k) \bbpsi_t^{(k-1)} + \beta_{k} \sum_{i\in\Omega_t}\epsilon_{i}\bbu_i[t-1]$
			\vspace{1.5mm}
%
			\EndFor
			\State \bf{return } $\bbpsi_t^{(K)}$
		\end{algorithmic}
	\end{center}
\end{algorithm}

\setcounter{subsec}{2}
\setcounter{algorithm}{0}

\begin{algorithm}[t]
	\caption{Newton method to obtain the sketch for the Probit model} \label{alg:psi:probit2}
	\begin{center}
		\begin{algorithmic} 
			\State\textbf{input:}~$\{y_{i,t},~i \in \Omega_{t}\}$, $\{\eta_j\}$, $\{\beta_k\}$, $\lambda$, $K$, $\sigma$, $\bbU[t-1]$
			
			\State\textbf{initialize:}~$\bbpsi_t^{(0)}$
			\For{$k = 1, 2, \ldots,K$}
			\State $a_{j,t-1}^i:={\sigma}^{-1}(\eta_{j}-\mathbf{u}_i^{\top}[t-1]\bbpsi_t^{(k-1)})$
			\vspace{-6mm}
			

		\State\begin{multline}
				\hspace{2mm}\theta_{i,t}:=\sum_{j=0}^{J-1}\mathcal{I}(y_{i,t}=s_j)\\
		\times\left[a_{j,t-1}^i{\rm \phi}(a^i_{j,t-1})-a_{j+1,t-1}^i{\rm \phi}(a_{j+1,t-1}^i)\right]\nonumber
		\end{multline}

		\State $f_{i,t}:=\sum_{j=0}^{J-1}\mathcal{I}(y_{i,t}=s_j)\left[{\rm \phi}(a_{j,t-1}^i)-{\rm \phi}(a_{j+1,t-1}^i)\right]$

		\State 	$w_{i,t}:=\sum_{j=0}^{J-1}\mathcal{I}(y_{i,t}=s_j)\left[{\rm Q}(a_{j,t-1}^i)-{\rm Q}(a_{j+1,t-1}^i)\right]$

		\vspace{1mm}

			\State $\nabla_{\bbpsi_t} g_t=-\sum_{i\in\Omega_t}\frac{f_{i,t}}{w_{i,t}}\bbu_i[t-1]+\lambda\bbpsi_t^{(k-1)}$
				\vspace{1mm}

			\State $\nabla^2_{\bbpsi_t} g_t=-\sum_{i\in\Omega_t}\left[\frac{f_{i,t}^2}{w_{i,t}^2}-\frac{\theta_{i,t}}{w_{i,t}}\right]\bbu_i[t-1]\bbu_i^\top[t-1]+\lambda\bbI$
			\vspace{1.5mm}
			\State 
			$\bbpsi_t^{(k)}=\bbpsi_t^{(k-1)}-\beta_{k}(\nabla^2_{\bbpsi_t} g_t)^{-1}\nabla_{\bbpsi_t} g_t$
			\vspace{1.5mm}
			
			\EndFor
			\State \bf{return } $\bbpsi_t^{(K)}$
		\end{algorithmic}
	\end{center}
\end{algorithm}

With the sketch $\{\bbpsi_{\tau}\}_{\tau=1}^t$ at hand, (S2) proceeds to update the subspace in (P3). This is however a daunting task since for the considered categorical models the regularized loss $g_t$ relates to the latent subspace $\bU$ in a complicated way (through functions  of the Gaussian pdf for the Probit and Tobit, and exponential functions for the Logit model), which precludes closed-form solutions. To bypass this computational hurdle, we will adopt an inexact solution of (P3). The basic idea leverages the empirical cost of (P3) to incorporate the information of the latest datum through a stochastic gradient descent iteration. In essence, at iteration (time) $t$ the old subspace estimate is updated by moving (with an appropriate step size) along the opposite gradient direction of $g_t$ incurred by the latest datum. All in all, this yields the recursion
\begin{align}
\mathbf{u}_i[t]=\mathbf{u}_i[t-1]-\mu_{t}\nabla_{\bbu_i} g_{t}\big(\{y_{i,t}\}_{i \in \Omega_t};\bbpsi_{t},\mathbf{U}[t-1]\big)\label{eq:subspace_update}
\end{align}
where $\mu_{t}$ is the step size that can vary across time.

For the Probit model, the gradient is simply obtained as 
\begin{align}
& \hspace{-.8cm}\nabla_{\bbu_i} g_t^{(\rm probit)} \big(\{y_{i,t}\}_{i \in \Omega_t};\bbpsi_{t},\mathbf{U}[t-1]\big) \nonumber \\&=-\frac{f(\bbu_i[t-1],\bbpsi_t)}{w(\bbu_i[t-1],\bbpsi_t)} \bbpsi_t+\frac{\lambda}{t}\bbu_i[t-1]\label{eq:gu:probit}
\end{align}
where the scalar functions $f$ and $w$ are given by
\begin{multline}
f(\bbu_i[t-1],\bbpsi_t) \nonumber \\ 
\hspace{-5mm} :=\sum_{j=0}^{J-1}\mathcal{I}(y_{i,t}=s_j)\left[{\rm \phi}(z_{j,t-1}^i)-{\rm \phi}(z_{j+1,t-1}^i)\right]\nonumber
 \end{multline}
 with~$z_{j,t-1}^i:={\sigma}^{-1}(\eta_{j}-\mathbf{u}_i^{\top}[t-1]\bbpsi_t)$, and
 \begin{multline}
w(\bbu_i[t-1],\bbpsi_t)\\:=\sum_{j=0}^{J-1}\mathcal{I}(y_{i,t}=s_j)
\left[{\rm Q}\left(z_{j,t-1}^i\right)-{\rm Q}(z_{j+1,t-1}^i)\right].\;\nonumber
\end{multline}

For the Tobit-I model, the gradient is expressed as
\begin{align}
&\nabla_{\bbu_i} g^{(\rm tobit-I)}_t\big(\{y_{i,t}\}_{i \in \Omega_t};\bbpsi_{t},\mathbf{U}[t-1]\big) \nonumber\\
& \hspace{0.cm}= \left\{
\begin{array}{ll}
-\frac{\left(y_{i,t}-\bbu_i^\top[t-1]\bbpsi_t\right)}{\sigma^2 }\bbpsi_t+\frac{\lambda}{t}\bbu_i[t-1],&	 y_{i,t}\in(\eta_l,\eta_u)\\
\frac{ {\rm \phi}(z_{u,t}^i)}{\sigma {\rm Q}(z_{u,t}^i)}\bbpsi_t+\frac{\lambda}{t}\bbu_i[t-1],&	  y_{i,t}=\eta_u\\
\frac{ {\rm \phi}(z_{l,t}^i)}{\sigma {\rm Q}(z_{l,t}^i)}\bbpsi_t+\frac{\lambda}{t}\bbu_i[t-1],&	 y_{i,t}=\eta_l\label{eq:gu:tobit}
\end{array}\right.
\end{align}
where $z_{u,t-1}^i:={\sigma}^{-1}\left(\eta_{u}-\bbu_i^{\top}[t-1]\bbpsi_t\right)$, and likewise for $z_{l,t-1}$. 
For the Tobit-II model, we have
\begin{align}
&\nabla_{\bbu_i} g^{(\rm tobit-II)}_t\big(\{y_{i,t}\}_{i \in \Omega_t};\bbpsi_{t},\mathbf{U}[t-1]\big) \nonumber\\
& \hspace{-0cm}= \left\{
\begin{array}{ll}
-\frac{{\rm \phi}(z^i_{l,t-1})-{\rm \phi}(z^i_{u,t-1})}{{\rm Q}\left(z^i_{l,t-1}\right)-{\rm Q}(z^i_{u,t-1})} \bbpsi_t+\frac{\lambda}{t}\bbu_i[t-1],&	 y_{i,t}\in(\eta_l,\eta_u)\\
-\frac{\left(y_{i,t}-\bbu_i^\top[t-1]\bbpsi_t\right)}{\sigma^2 }\bbpsi_t+\frac{\lambda}{t}\bbu_i[t-1],&	 y_{i,t}=\eta_l, {\rm or}~\eta_u.\label{eq:gu:tobit2}
\end{array}\right.
\end{align}
Finally, one can arrive at the gradient of the binary Logit model that is given by
\begin{align}
&\nabla_{\bbu_i} g^{(\rm logit)}_t\big(\{y_{i,t}\}_{i \in \Omega_t};\bbpsi_{t},\mathbf{U}[t-1]\big)\nonumber\\
=&\frac{(2y_{i,t}-1)\exp\{(2y_{i,t}-1)\bbu_i^{\top}[t-1]\bbpsi_t\}}{1+\exp\{(2y_{i,t}-1)\bbu_i^{\top}[t-1]\bbpsi_t\}}\bbpsi_t+\frac{\lambda}{t}\bbu_i[t-1].
\end{align}

\vspace{-2mm}

The subspace update \eqref{eq:subspace_update} amounts to exactly solving a first-order approximation of the cost in (P3). The overall procedure is summarized in Algorithm \ref{algo:online} only for the Probit model, but it also applies for the Tobit and Logit models with obvious modifications for the gradient correction terms.

\noindent\textbf{Remark 1 (Computational cost):}~The subspace update in Algorithm \ref{algo:online} is parallelizable across columns ($d$), and can be efficiently implemented via GPUs. The major complexity emanates from running the iterative Algorithm~\ref{alg:psi:probit} or \ref{alg:psi:probit2}, for obtaining $\bpsi_t$. Fixing the maximum number of inner iterations to $K$, this demands $\mathcal{O}(Kd^2D)$ operations for Algorithm~\ref{alg:psi:probit2}, and $\mathcal{O}(KdD)$ operations for Algorithm~\ref{alg:psi:probit}. Our empirical observations suggest that even an inexact solution of (S1) obtained by running Algorithm \ref{alg:psi:probit2} with a few iterations $K$ suffices for  Algorithm \ref{algo:online} to converge. The remaining operations entail multiplications and additions of order $\mathcal{O}(D)$. The overall cost of the Algorithm \ref{algo:online} per iteration is $\mathcal{O}(Kd^2D)$, which is affordable since $d$ is generally small.

\section{Learning the quantizer} \label{sec:learn_quan}
The Probit model discussed in the previous sections requires quantization thresholds $\{\eta_j\}_{j=0}^{J-1}$ to be available. These thresholds however add degrees of freedom, which can enhance the predictive power of the Probit based approach to modeling categorical data. While one can derive the general multibit case, to simplify exposition, consider the binary case with a single threshold $\eta$ that is assumed fixed over time. With this in mind, \eqref{eq:model:probit} boils down to 
\begin{align}
\label{probit:th}
y_{i,t}= {\rm sign} (\bbu_i^{\top} \bbpsi_t + v_{i,t}-\eta).
\end{align}
To sketch big categorical data obeying \eqref{probit:th}, both $\bbu_i$ and $\eta$ must be selected jointly. An estimate of these parameters can be found by jointly maximizing the rank-regularized likelihood in (P2), where the per-entry log-likelihood is now replaced by
\begin{align}
\log\ell_{\rm probit}(y_{i,t};\bbu_i,\bbpsi_{t},\eta)&
=\frac{1+y_{i,t}}{2}  \log   {\rm Q}\left( \frac{\eta-\mathbf{u}_i^{\top}\bbpsi_{t}}{\sigma} \right)\nonumber\\
&+\frac{1-y_{i,t}}{2} \log {\rm Q}\left(\frac{\mathbf{u}_i^{\top}\bbpsi_{t}-\eta}{\sigma} \right) \;. \nonumber
\end{align}
Accordingly, the updates for $\{\bbu_i\}$ and $\eta$ are obtained by applying stochastic gradient descent to the empirical loss in (P3). 

The sketch and subspace updates are similar to \eqref{eq:online1} and \eqref{eq:subspace_update}, while $\eta$ is updated as
\begin{align}
\eta[t]=\eta[t-1]-\gamma_{t}\nabla_{\eta} g_{t}\big(\{y_{i,t}\}_{i \in \Omega_t};\bbpsi_{t},\mathbf{U}[t],\eta\big)\;\label{eq:online2}
\end{align}
where the gradient with respect to $\eta$ is readily expressed as
\begin{align}
\nabla_{\eta} g_t=-\sum_{i\in\Omega_t}\zeta_{i,t-1}
\end{align}
where
$
\zeta_{i,t-1}:=-b_{i,t-1} \sigma^{-1}{{\rm \phi}(b_{i,t-1})}/{{\rm Q}\left(b_{i,t-1}\right)}
$,	
and
$
b_{i,t-1}:=\sigma^{-1}y_{i,t}\left( \eta[t-1]-\mathbf{u}_i^{\top}[t-1]\bbpsi_t\right).
$

Albeit more complex, analogous updates are possible for the multibit Probit, and likewise for designing the quantizer when Tobit and Logit models are adopted.

\vspace{0.1cm}

\section{Performance Analysis} \label{sec:perf_analysis}
This section establishes convergence of the first-order iterates in Algorithm \ref{algo:online} for the considered categorical models, namely Probit, Tobit, and Logit. Both asymptotic and non-asymptotic analyses for infinite and finite data streams are considered. The asymptotic analysis relies heavily on quasi-martingale sequences~\cite{mairalonlinelearning}, while for non-asymptotic analysis we draw from regret metric advances in online learning~\cite{banerjee12,ZC14,shalev2011online}.

\subsection{Asymptotic convergence analysis}
For {\it infinite} data streams, convergence analysis of our categorical subspace learning schemes is inspired by \cite{mairalonlinelearning}, and our precursors in \cite{mardani2015subspace} and \cite{MMG13JSTSP}. In order to render analysis tractable, the following assumptions are adopted.

  {\it (as1) The  data streams $\{\by_t\}_{t=1}^{\infty}$ and sampling patterns $\{\Omega_t\}_{t=1}^{\infty}$ form an i.i.d. process; and}
	
{\it (as2) the subspace sequence $\{\bU[t]\}$ lies in a compact set.}

To begin, rewrite the rank-regularized empirical cost in (P3) as
\begin{align}
\min_{\bU \in \mathbbm{R}^{D \times d}}~~~C_t(\bU):=\frac{1}{t} \sum_{\tau=1}^{t}~g_{\tau}(\bpsi_{\tau},\bU).  \label{eq:avg_cost_Ct}
\end{align}
As argued earlier in Section \ref{sec:online}, minimization of \eqref{eq:avg_cost_Ct} becomes increasingly complex computationally as $t$ grows. The subspace $\bU[t]$ is estimated by the stochastic gradient-descent (SGD) iteration with an appropriate step size. SGD iterations can be seen as minimizing the {\it approximate} cost
\begin{align}
\label{C:tilde}
\check{C}_t(\bU) = \frac{1}{t} \sum_{\tau=1}^{t} \check{g}_{\tau}(\bpsi_{\tau},\bU) 
\end{align}
where $\check{g}_t$ is a quadratic upperbound for $g_t(\cdot)$ based on the second-order Taylor approximation around the latest subspace estimate $\bU[t-1]$; that is
{\small
\begin{multline}
\check{g}_t(\bpsi_t,\bU) = g_t(\bpsi_t,\bU[t-1]) + \langle \nabla_{\bU} g_t(\bpsi_t,\bU[t-1]), \bU-\bU[t-1] \rangle\\ + \frac{\alpha_t}{2} \|\bU-\bU[t-1]\|_F^2
\end{multline}}
with $\alpha_t \geq \|\nabla^2_{\bbU} g_t(\bpsi_t,\bU[t-1])\|$. It is useful to recognize that the quadratic surrogate $\check{g}_t(\cdot)$ is a tight approximation for $g_t$, since (i) it is an upperbound, i.e., $\check{g}_t(\bpsi_t,\bU) \geq g_t(\bpsi_t,\bU),~\forall \bU$; and (ii), it is locally tight, i.e., $\check{g}_t(\bpsi_t,\bU[t-1]) = g_t(\bpsi_t,\bU[t-1])$, with (iii) locally tight gradient, i.e., $\nabla \check{g}_t(\bpsi_t,\bU[t-1]) = \nabla g_t(\bpsi_t,\bU[t-1])$. Furthermore, $g_t$ is smooth as asserted next.


\begin{lemma}
	Under (as2), upon defining $\delta_1:=\Delta/\sigma^2$, $\delta_2:=(\Delta^2/\sigma^2+1)/\sigma^2$, and $\Delta:=\eta_{J-1}-\eta_0$, for the gradient and Hessian of the per-entry loss for the Probit model, it holds that 
	\begin{align}
	\|\nabla_{\bbu_i} g^{\rm (probit)}_t\big(\bbpsi_{t},\mathbf{U}\big)\|_2&\leq \delta_1 \|\bbpsi_t\|_2+\frac{\lambda}{t}\|\bbu_i\|_2	\\
			\|\nabla_{\bbu_i}^2 g^{\rm (probit)}_t\big(\bbpsi_{t},\mathbf{U}\big)\|&\leq \delta_2\|\bbpsi_t\|_2^2+\frac{\lambda }{t}
	\end{align}
	and consequently the per-entry cost $g^{\rm (probit)}_t\big(\bbpsi_{t},\mathbf{U}\big)$, and $\nabla g^{\rm (probit)}_t\big(\bbpsi_{t},\mathbf{U}\big)$ are Lipschitz continuous. \label{lemma1}
\end{lemma}

\begin{IEEEproof}
	See the Appendix. 
\end{IEEEproof}

The convergence of subspace iterates can then be established following the machinery developed in \cite{mairalonlinelearning}. In the sequel, technical details are skipped due to space limitations, but they follow  arguments similar to those in \cite{mardani2015subspace}. The proof sketch entails the following two main steps.

\textbf{(Step1)} The approximate cost $\check{C}_t(\bU[t])$ asymptotically converges to $C_t(\bU[t])$, i.e., $\lim_{t \rightarrow \infty} |C_t(\bU[t])-\check{C}_t(\bU[t])|=0$. The convergence follows the quasi-martingale property of $\{\check{C}_t\}$ in the almost sure (a.s.) sense owing to the tightness of the surrogate function $\check{g}_t$.       

\textbf{(Step2)} Due to the regularity of $g_t$, asymptotic convergence of $\{C_t(\bU[t])-\check{C}_t(\bU[t])\} \rightarrow 0$ implies convergence of the associated gradient sequence, namely $\{\nabla C_t(\bU[t])-\nabla \check{C}_t(\bU[t])\} \rightarrow 0$, which ultimately leads to $\nabla C_t(\bU[t]) \rightarrow \mathbf{0}$.

The projection coefficients $\bpsi_t$ can be solved exactly using Newton iterations due to the convexity of $g_t(\bpsi_t,\bU[t-1])$, when the subspace is frozen at $\bU[t-1]$. This is formalized in the next lemma.

\begin{lemma}\label{lemma2}
	Under the Probit, Tobit-II, and Logit models, the per-entry  regualrized-loss $g_t(\bpsi,\bU)$ is bi-convex for the block variables $\bpsi$ and $\bu_i$. 
\end{lemma}

\begin{IEEEproof}
See the Appendix.
\end{IEEEproof}

All in all, combining the previous arguments with Lemmas 1 and 2, the asymptotic convergence claim for the iterations of Algorithm \ref{algo:online} can be asserted as follows.

\begin{proposition} \label{prop:prop_1}
Suppose (as1)-(as2) hold, and choose the step-size sequence $\{\mu_t=1/\bar{\alpha}_t\}$ where $\bar{\alpha}_t \geq ct$, and $\delta_2 \|\bpsi_t\|^2 + \lambda/t \leq \alpha_t \leq c'$ for constants $c,c'>0$, and $\delta_2$ as in Lemma \ref{lemma1}. Then, the subspace sequence $\{\bU[t]\}$ satisfies $\lim_{t \rightarrow \infty}\nabla_{\bU} C_t(\bU[t]) = \mathbf{0}$, which means that the subspace iterates asymptotically converge to the stationary-point set of the batch ML estimator (P1).
\end{proposition}

%
%
%

\subsection{Regret analysis}
For {\it finite} data streams, we will rely on the unsupervised formulation of regret analysis to assess the performance of online iterates, in terms of interpolating misses and denoising the available categorical data. Regret analysis was originally introduced for the online supervised learning scenario~\cite{shalev2011online}, where the ground-truth label is revealed after prediction to incur a loss whose gradient is used to guide the learning. In the considered unsupervised sketching task however, the true labels are not revealed, which challenges regret analysis. Unsupervised variations of regret have been lately introduced to deal with online dictionary learning \cite{banerjee12}, and sequential logistic PCA \cite{ZC14}.

Prompted by the alternating nature of iterations, we adopt a variant of the unsupervised regret to assess the goodness of online subspace estimates in representing the partially available data. Specifically, at iteration $t$, we use the previous update $\bU[t-1]$ to span the recent partial data, namely, $y_{i,t},~i \in \Omega_t$. With $g_t(\bpsi_t,\bU[t-1])$ being the loss incurred by the estimate $\bU[t-1]$ for predicting the $t$-th datum, the cumulative online loss for a stream of size $T$ is given by
\begin{align}
\bar{C}_T:= \frac{1}{T}\sum_{\tau=1}^T g_{\tau}(\bbpsi_{\tau},\bbU[\tau-1]).\label{eq:c:bar}
\end{align}
Further, we will assess the cost of the last estimate $\bU[T]$ using
\begin{align}
\hat{C}_T= \frac{1}{T}\sum_{\tau=1}^T g_{\tau}(\bbpsi_{\tau},\bbU[T]).\label{eq:avg_online_loss}
\end{align}
Comparing the losses in \eqref{C:tilde}, \eqref{eq:c:bar}, and \eqref{eq:avg_online_loss}, with $C_T:=\min_{\bU} C_T(\bU)$, it clearly holds that  
\begin{align}
\check{C}_T\geq\hat{C}_T \geq \bar{C}_T \geq C_T .\label{eq:cost_order}
\end{align}
Accordingly, for the sequence $\{\bU[t]\}_{t=1}^T$, define the online regret 
\begin{align}
\mathcal{R}_T:= \hat{C}_T - \bar{C}_T.    \label{def:regret}
\end{align}
Our next goal is to investigate the convergence rate of the sequence $\{\mathcal{R}_T\}$ to zero as $T$ grows. This is important particularly because it is known from Proposition \ref{prop:prop_1} that $|\check{C}_t - C_t| \rightarrow 0$ as $t \rightarrow \infty$, and as a result $|\bar{C}_t - C_t| \rightarrow 0$ (cf.~\eqref{eq:cost_order}). Due to the nonconvexity of the online subspace iterates, it is challenging to directly analyze how fast the online cumulative loss $\bar{C}_t$ approaches the optimal batch cost $C_t$. Instead, we will investigate whether $\hat{C}_t$ converges to $\bar{C}_t$.

In the sequel, to derive regret bounds we focus on the Probit model. However, the same analysis caries over to develop regret bounds for the Tobit and Logit models too.

\begin{proposition}\label{pro:regret}
	If $\{\bbU[t]\}$ and $\{\bpsi_t\}$ are uniformly bounded, i.e.,~ $\|\bbU[t]\|_F\leq B_u$, and $\|\bbpsi_t\|_2\leq B_{\psi}$ for constants $B_{u}, B_{\psi} >0$, choosing a constant step size $\mu_t=\mu$, leads to a bounded regret as
	\begin{align}
	\mathcal{R}_T\leq&  \frac{B^2(\ln(T)+1)^2}{2\mu T} +\frac{5 B^2}{6\mu T}\nonumber
	\end{align}
	where $B:={( \lambda B_u+\delta_1 B_{\psi})}/{\rho}$ is a constant not dependent of $T$, $\delta_1$ as in Lemma \ref{lemma1}, and $\rho$ denotes the strong convexity constant on $\bar{C}_T$. 
\end{proposition}

\noindent\textbf{Remark 3~[Subpace Projection]:}~Instead of assuming bounded subspace iterates, namely $\|\bbU[t]\|_F\leq B_u$, one can alternatively introduce an additional projection onto the $B_u$-ball given by $\{\bbU|~\|\bbU\|_F\leq B_u\}$. This additional projection does not alter the asymptotic convergence result in Proposition \ref{pro:regret} due to the non-expansiveness of the projection operator.

To place Proposition~\ref{pro:regret} in context, relevant regret analyses have been carried out for the dictionary learning~\cite{banerjee12}, and the sequential logistic PCA~\cite{ZC14}. Different from our scheme, \cite{banerjee12} deals with overcomplete dictionary updates with sparsity-regularized projection coefficients, and assumes that the estimation error is uniformly bounded. The regret bound obtained in \cite{ZC14} for logistic PCA also assumes no absent data entires, and it is relatively loose since the regret does not vanish as $T \rightarrow \infty$.

The  proof technique of Proposition \ref{pro:regret} relies on the following lemma, which asserts that the distance between successive subspace estimates vanishes as fast as $o(1/t)$, a property that will be instrumental to establish sub-linearity of the regret later. 

\begin{lemma}\cite{MMG13JSTSP}  \label{lem:lemma3}
	Under (as2), it holds that 
	\begin{align}
	\|\bbU[t]-\bbU[t-1]\|_F\leq \frac{B}{t}\nonumber
	\end{align}
	for some constant $B:={ (\lambda B_u+\delta_1 B_{\psi})}/{\rho}$, where $\rho$ denotes the strong convexity constant of $\bar{C}_t$. 
\end{lemma}

\begin{IEEEproof}
See the Appendix.
\end{IEEEproof}

Toward bounding the regret, consider the difference of the iterates (cf. \eqref{eq:subspace_update}) 
\begin{align}
\bbU[t]-\bbU[t-1]=-\mu_t\nabla_{\bbU}g_t(\bbpsi_t,\bbU[t-1]).
\end{align}
Taking the Frobenius norm on both sides yields
\begin{align}
\label{eq:boundu}
	\|\bbU[t]-\bbU{t-1}\|_F=\mu_t\|\nabla_{\bbU}g_t(\bbpsi_t,\bbU[t-1])\|_F
\end{align}
and after appealing to Lemma \ref{lem:lemma3}, we arrive at
\begin{align}
\label{bound:grad:u}
	\|\nabla_{\bbU}g_t(\bbpsi_t,\bbU[t-1])\|_F\leq\frac{B}{\mu_t t}.
\end{align}
On the other hand, it is easy to verify that (cf. \eqref{eq:boundu})
\begin{align}
&\|\bbU[t]-\bbU[T]\|_F^2\nonumber\\
&=\|\bbU[t-1]-\bbU[T]+\bbU[t]-\bbU[t-1]\|_F^2\nonumber\\
&=\|\bbU[t-1]-\bbU[T]\|^2_F+\mu_t^2\|\nabla_{\bbU}g_t(\bbpsi_t,\bbU[t-1])\|_F^2\nonumber\\
&\hspace{5mm}-2\mu_t \langle \bbU[t]-\bbU[T], \nabla_{\bbU}g_t(\bpsi_t,\bU[t-1])\rangle\nonumber
\end{align}
which after re-arranging yields
\begin{align}
&\langle \bbU[t]-\bbU[T], \nabla_{\bbU}g_t(\bpsi_t,\bU[t-1])\rangle=\frac{\|\bbU[t-1]-\bbU[T]\|^2_F}{2\mu_t}\nonumber\\
&\hspace{0.5cm}+\frac{\mu_t\|\nabla_{\bbU}g_t(\bbpsi_t,\bbU[t-1])\|_F^2}{2}-\frac{\|\bbU[t]-\bbU[T]\|^2_F}{2\mu_t}.\label{eq:1}
\end{align}
Thanks to the separability of $g_t$, along with its convexity (cf. Lemma~\ref{lemma2}), one can establish the inequality
\begin{align}
& g_t(\bbpsi_t,\bbU[T])- g_t(\bbpsi_t,\bbU[t-1])\nonumber\\
&\geq \langle \bbU[T]-\bbU[t-1], \nabla_{\bbU}g_t(\bbpsi_t,\bbU[t-1])\rangle.\label{eq:cnvx}
\end{align}
Using \eqref{eq:cnvx}, this yields the following upper bound
\begin{align}
& T\left[\bar{C}_T-\hat{C}_T\right]\nonumber\\
&=\sum_{t=1}^T\left[ {g}_t(\bbpsi_t,\bbU[t-1])- g_t(\bbpsi_t,\bbU[T])\right]\nonumber\\
&\leq \sum_{t=1}^{T} \langle \bbU[t-1]-\bbU[T], \nabla_{\bbU}g_t(\bbpsi_t,\bbU[t-1])\rangle.\label{ineq:3}
\end{align}
Substituting \eqref{eq:1} into \eqref{ineq:3}, and combining with \eqref{bound:grad:u}, leads to
\begin{align}
& T\left[\bar{C}_T-\hat{C}_T\right]\leq  \frac{\|\bbU[0]-\bbU[T]\|_F^2}{2\mu_1}\nonumber\\
+& \sum_{t=1}^{T}\left(\frac{1}{2\mu_{t+1}}-\frac{1}{2\mu_t}\right)\|\bbU[t-1]-\bbU[T]\|_F^2+\frac{B^2}{2}\sum_{t=1}^T\frac{1}{\mu_t t^2}.\label{ineq}
\end{align}
%
%
Regarding the first term in the right hand side of \eqref{ineq}, it can be further bounded by
\begin{align}
&\frac{\|\bbU[0]-\bbU[T]\|_F^2}{2\mu_1}\nonumber\\
&=\frac{1}{2\mu_1}\|\bbU[0]-\bbU[1]+\bbU[1]-\bbU[2]+\cdots+\bbU[T-1]-\bbU[T]\|_F^2\nonumber\\
&\leq \frac{1}{2\mu_1}(\|\bbU[0]-\bbU[1]\|_F+\cdots+\|\bbU[T-1]-\bbU[T]\|_F)^2\nonumber\\
&= \frac{1}{2\mu_1}\left(\sum_{t=1}^T \|\bbU[t]-\bbU[t-1]\|_F\right)^2\nonumber\\
&\leq \frac{1}{2\mu_1}\left(\sum_{t=1}^T \frac{B}{t}\right)^2\nonumber\\
&\leq \frac{B^2}{2\mu_1} (\ln(T)+1)^2
\end{align}
where the first inequality follows from the triangle inequality, while the last two inequalities are due to Lemma \ref{lem:lemma3} and the property of harmonic series, respectively. 
Upon choosing a constant step size $\mu_t=\mu$, the last term in \eqref{ineq} can be bounded by \cite{ayoub1974euler}
\begin{align}
	\frac{B^2}{2\mu}\sum_{t=1}^T\frac{1}{ t^2}\leq \frac{5 B^2}{6\mu}
\end{align}
and after some algebra one arrives at
\begin{align}
\bar{C}_T-\hat{C}_T \leq  \frac{B^2(\ln(T)+1)^2}{2\mu T} +\frac{5 B^2}{6\mu T}\nonumber
\end{align}
which completes the proof of Proposition \ref{pro:regret}.

\section {Numerical Tests}\label{sec:test}
Performance of the novel online categorical subspace learning schemes is assessed in this section via simulated tests on both synthetic and real datasets. The real datasets are: (D1) the chess-game dataset ``King-Rook versus King-Pawn'' dataset~\cite{D00}; and (D2) the user-movie rating dataset ``MovieLens100K''~\cite{MAL03}.



\subsection{Synthetic data}\label{subsec:syn}

Synthetic categorical data $\{\by_t\}_{t=1}^T$ with $D=25$ across $T=5,000$ time instants are generated after quantizing the real-valued process $\{\bx_t=\bU\bbpsi_t\}_{t=1}^T$ to the alphabet $\cS:=\{1,\ldots,5\}$. The underlying low-dimensional sketch is drawn equiprobably from two populations, namely $\psi_{i,t} \sim \cN(-1,0.04)$ for the first class; and $\psi_{i,t} \sim \cN(+1,0.04)$ for the second class. Matrix $\bbU\in\mathbb{R}^{D\times d}$ is generated with entries drawn from the standardized normal distribution. Uniform quantizer is adopted with thresholds $\eta_j:=\frac{-J+1+2j}{J-1}x_{\max},~j=0,1,\ldots,J-1$, where $x_{\max}$ denotes the maximum absolute entry of $\bx_t$. To simulate the missing entries, a subset of entries are dropped uniformly at random with probability $1-p$.

%

Throughout the tests a constant step size $\mu_{t}=0.01$ is adopted for the subspace update, and the rank controlling parameter is set to $\lambda=0.1$. The results are averaged over $100$ independent trials.

\begin{figure}[h]
	\centering
	\includegraphics[width=8cm]{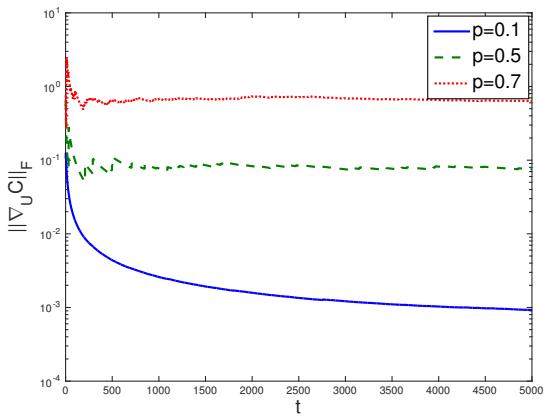}
	\caption{Empirical gradient-norm of (P3) versus time for synthetic data under variable $\%$ of misses ($1-p$).}\label{fig:gr}
\end{figure}

%
%
\begin{figure*}[t]\label{fig:sketch}
\begin{minipage}[b]{.33\textwidth}
\centering
\includegraphics[width=6cm]{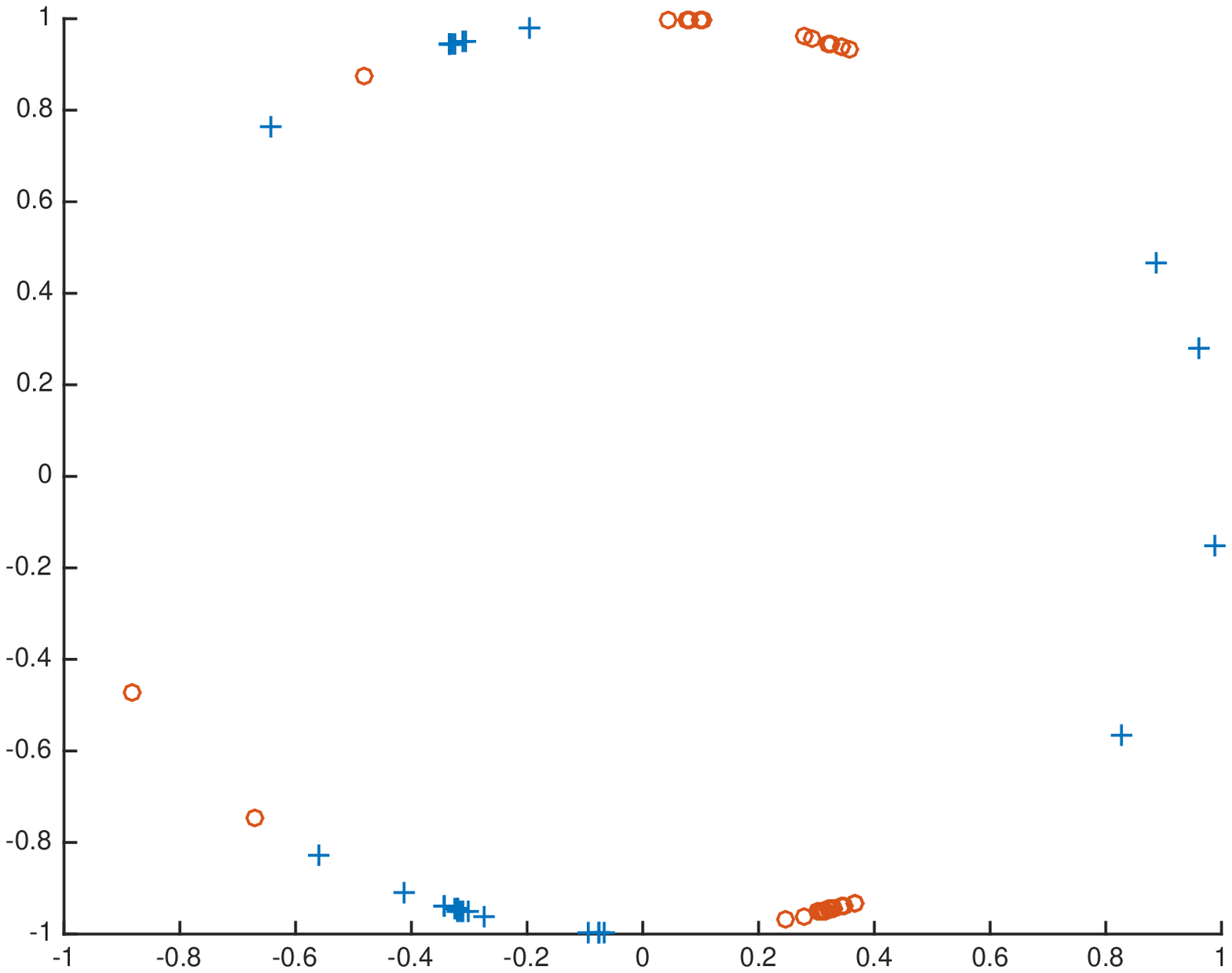}
\centerline{(a)}
\end{minipage}
\begin{minipage}[b]{.33\textwidth}
\centering
\includegraphics[width=6cm]{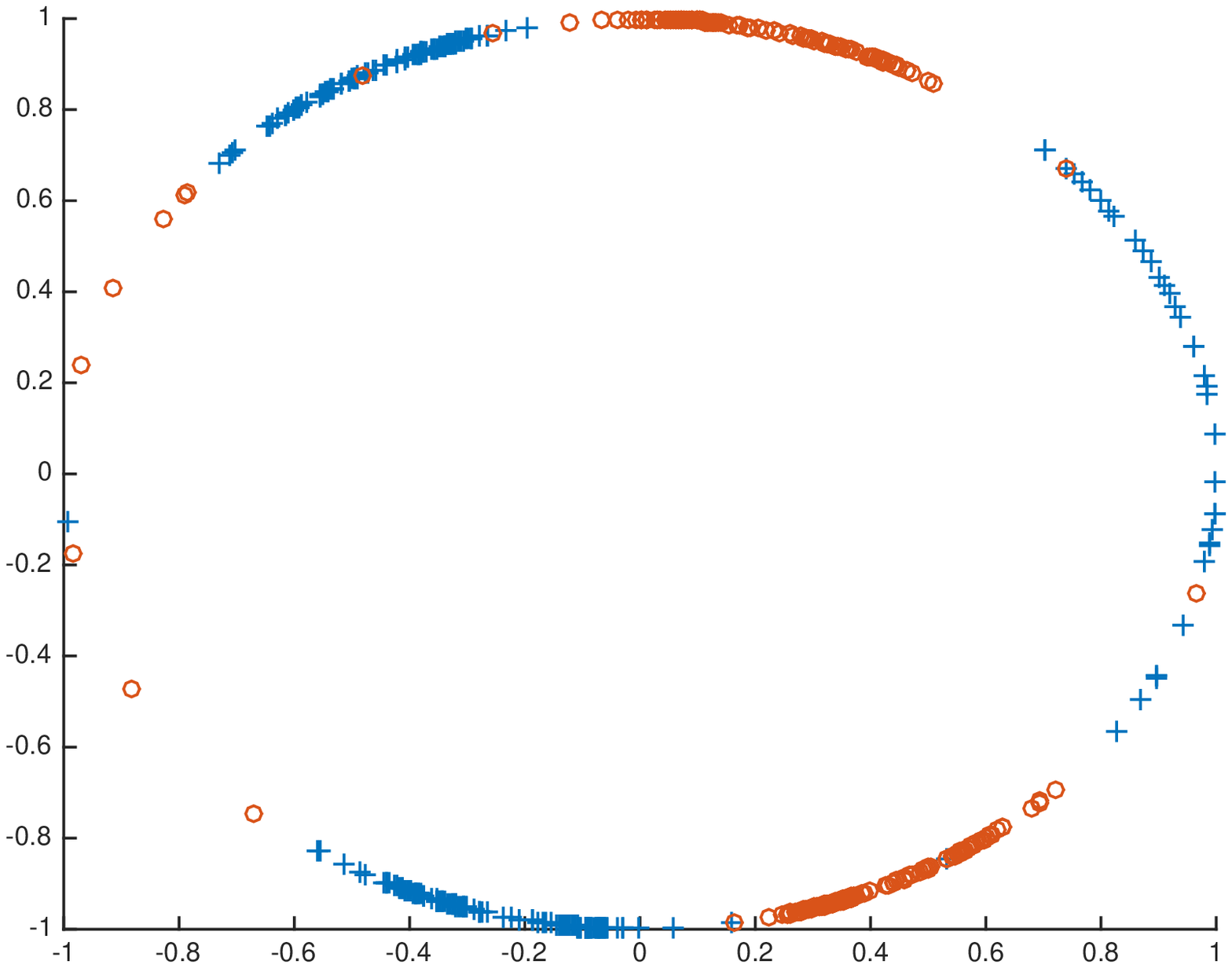}
\centerline{(c)}
\end{minipage}
\begin{minipage}[b]{.33\textwidth}
\centering
\includegraphics[width=6cm]{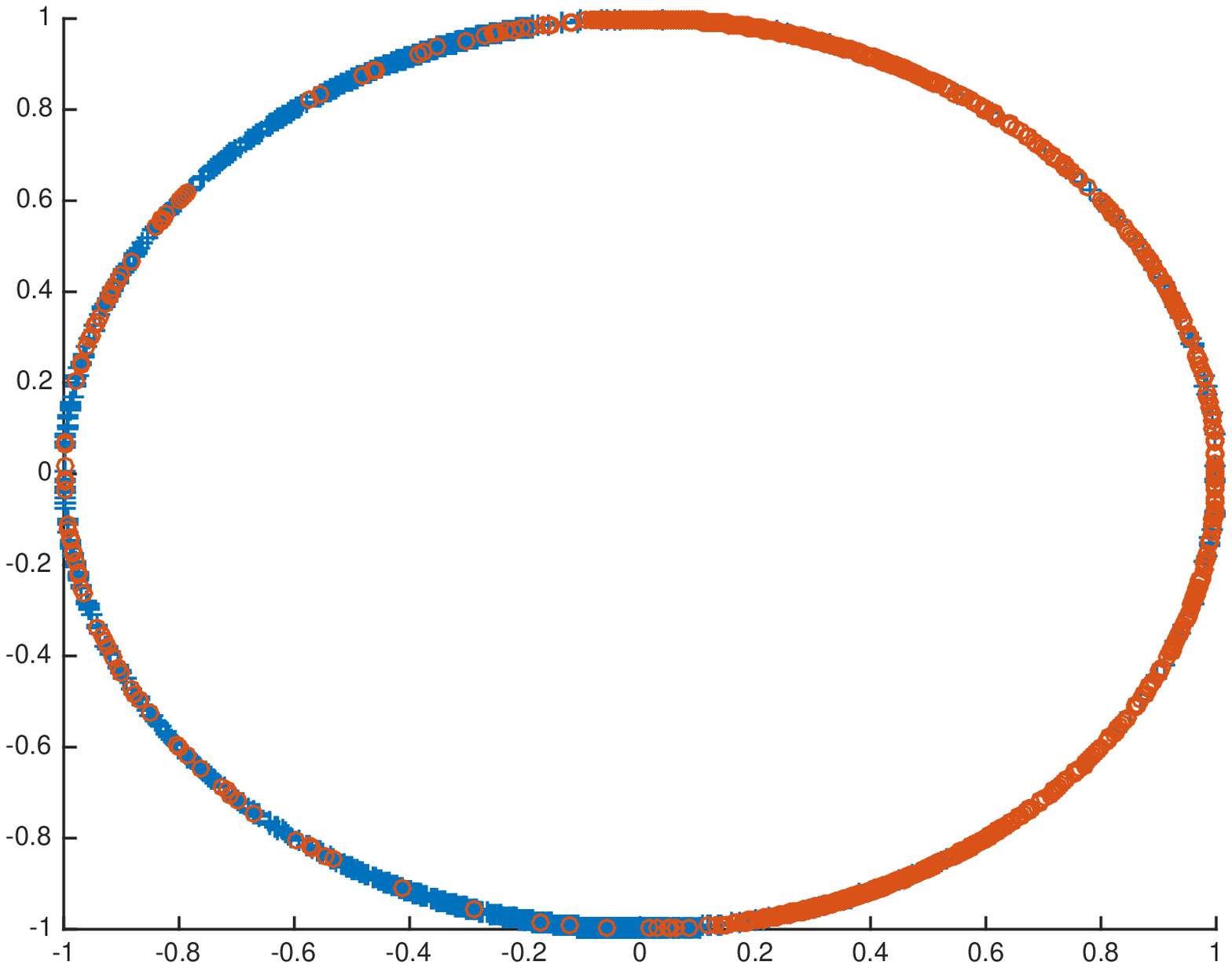}
\centerline{(c)}
\end{minipage}
\vspace{1mm}
 \caption{Time evolution of the sketch obtained using the Probit model at (a) $t=30$; (b) $t=300$; and, (c) $t=3,000$, when $d=2$, and $D=5$.} 
 \label{fig:skeches}
\end{figure*}

\begin{table}[h]
\begin{center}
		\begin{tabular}{ c| c | c }
			\hline
			\hline
			& \multicolumn{2}{c}{online CSL}\\ \hline
			$p$  & runtime (sec) & classification error ($\%$)  \\ \hline
			0.1 & 3.0333  &42.17 \\ \hline
			0.3 & 2.5925 &17.20\\ \hline
			0.5 & 2.7029  & 4.76\\ \hline
			0.7& 2.8967 &2.18\\
			\hline
			\hline
			& \multicolumn{2}{c}{MM \cite{LHH10}}\\ \hline
			0.1 & 8.1267  & 43.86\\ \hline
			0.3  & 6.4973 & 32.62\\ \hline
			0.5 &7.5499 & 17.01\\ \hline
			0.7& 9.2077 & 8.31\\
			\hline
			\hline
		\end{tabular}
		\caption{Runtime (seconds) and classification error comparison of the novel scheme against the batch MM~\cite{LHH10} for synthetic data under variable fraction of misses $1-p$.}\label{tab:syn}
	\end{center}
\end{table}

Convergence of Algorithm \ref{algo:online} under various percentages of missing data is demonstrated in Fig. \ref{fig:gr} depicting the empirical gradient-norm (w.r.t. $\bbU$) of (P3) over time. It is evident that after about $1,200$ iterations, the online algorithm with random initialization attains a stationary point of (P2). To highlight the merits of the novel scheme, the batch majorization-minimization (MM) scheme of \cite{LHH10} is also implemented. In essence, MM relies on the Logit model with binary data ($J=2$), and thus one needs first to obtain binary categorical data to make it operational. Setting $d=8$, the low-dimensional sketch returned by both algorithms is used to classify the data using a linear SVM classifier. The resulting runtime as well as the classification error (fraction of miss-classified data) for our scheme and MM are listed in Table \ref{tab:syn} for a fraction of absent entries. It is apparent that our online scheme exhibits considerable advantage in runtime and accuracy over the batch MM scheme, while also offering real-time sketching and classification of data `on the fly.'

To further illustrate the operation of real-time sketching, we tested the binary quantization model $y_{i,t}=\text{sign}(\bbu_i^\top\bbpsi_t+v_{i,t})$ with $v_{i,t}\sim\mathcal{N}(0,0.01)$, and $\bbU$ generated from the standardized normal distribution. The two-dimensional sketch $\bpsi_t$ is drawn equiprobably as $[1,1]^\top$ for the first class, and $[-1,-1]^\top$ for the second class. The sketch evolution is depicted in Fig. \ref{fig:skeches} at different time instants $t=30,300,3000$, where it is evident that as more data arrive, the latent subspace is learnt more accurately, and consequently the data points are assigned to the correct classes.

\begin{figure*}[t]
\centering
\hspace{0mm}\begin{minipage}[b]{.52\textwidth}
\centering
\includegraphics[width=8.8cm]{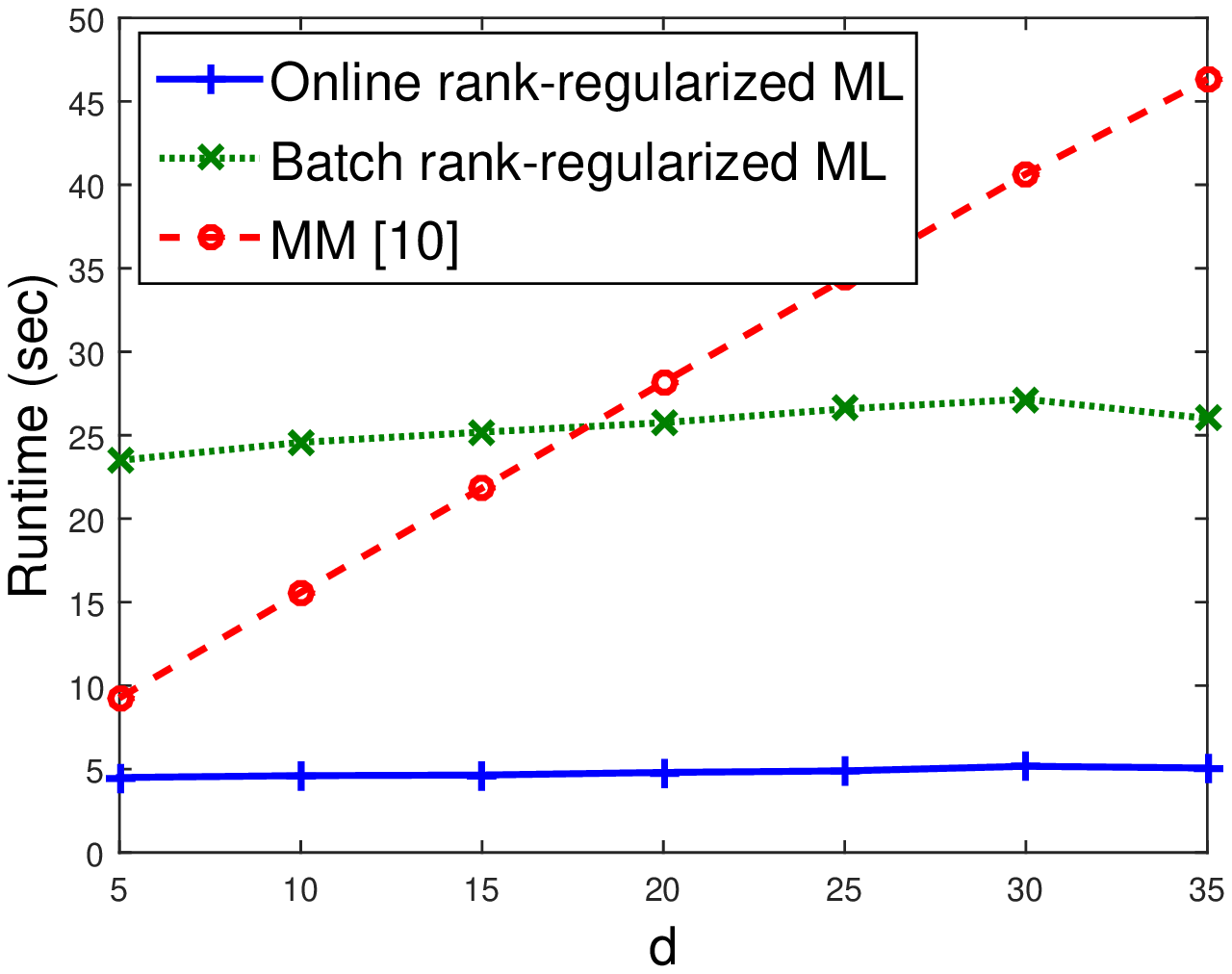}
\centerline{(a)}
\end{minipage}
\hspace{-10mm}\begin{minipage}[b]{.52\textwidth}
\centering
\includegraphics[width=8.8cm]{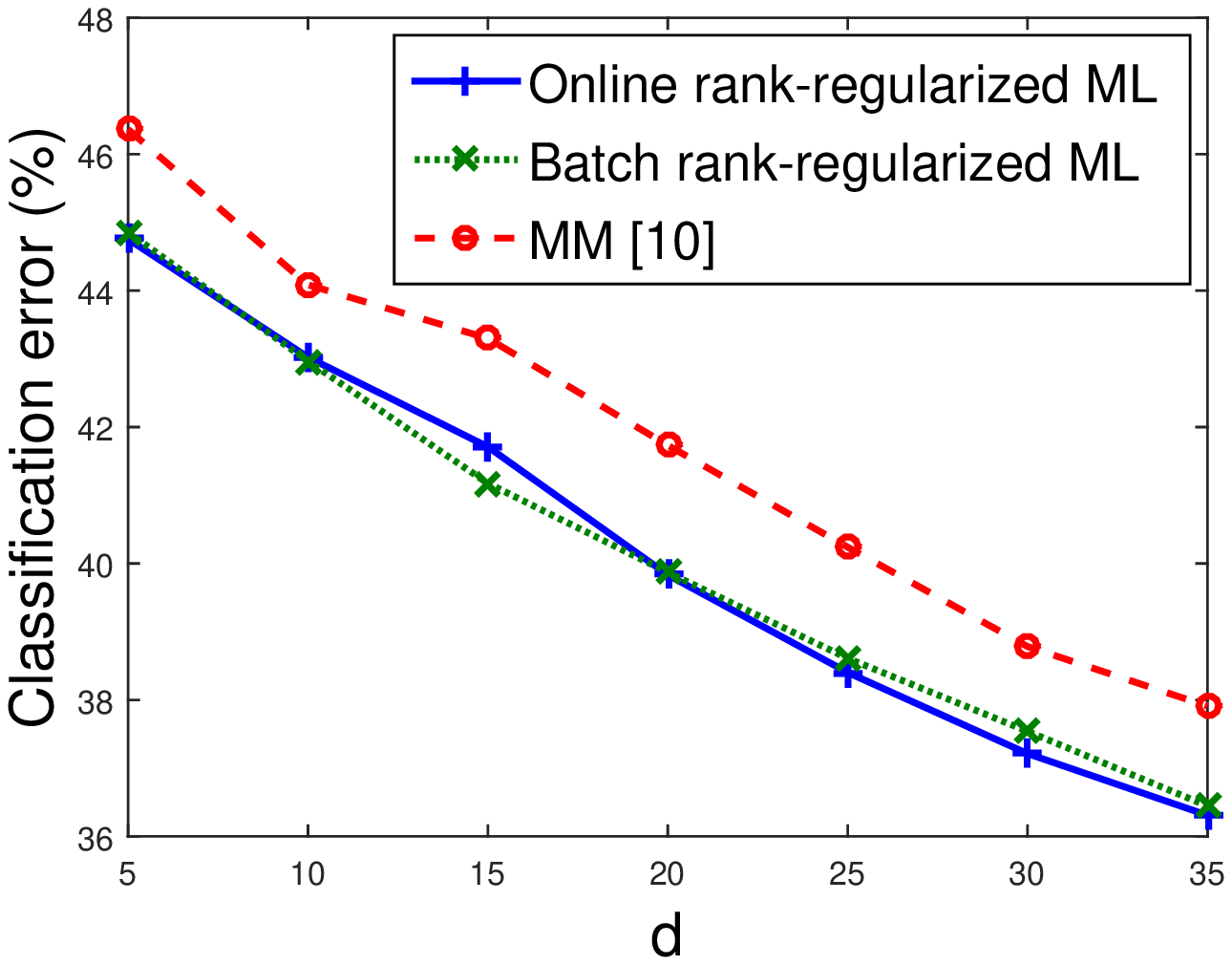}
\centerline{(b)}
\end{minipage}
\vspace{1mm}
 \caption{Runtime (left) and LS classification error (right) of the novel CSL scheme versus the MM scheme~\cite{LHH10} for the ``King-Rook versus King-Pawn'' dataset under variable dimension $d$ when $p=0.1$, $D=35$, and $T=3,196$.} 
 \label{fig:chess}
\end{figure*}

\subsection{Classification of chess games}
In this experiment, we considered the chess-game dataset ``King-Rook versus King-Pawn'' acquired across $T=3,196$ scenarios, each with $D=35$ binary ($J=2$) data signifying nominal attributes. The online sketch returned by Algorithm \ref{algo:online} is used to group games in two classes, namely ``white-can-win'' and ``white-cannot-win,'' upon averaging the classification outcomes over $100$ independent runs. As it is evident from Fig. \ref{fig:chess}(a) with $90\%$ random misses ($p=0.1$), our novel approach achieves considerable runtime advantage over the MM scheme for sketching the partial data, especially when the dimension of the latent subspace is in the order of a few dozens. With the low-dimensional sketch at hand, LS classification~\cite{bishop06book} is performed, and the resultant error is plotted in Fig. \ref{fig:chess}(b) under different compression ratios. Our novel CSL-based scheme consistently improves the classification accuracy by about $5\%$ relative to MM, indicating that the adopted model better matches the considered real-world dataset. Note that our scheme only relies on a {\it single} pass over the dataset.

%
%

\subsection{Interpolation of MovieLens dataset}
The MovieLens dataset~(D2) is considered to evaluate the interpolation capability of the novel CSL scheme. This dataset contains $100,000$ discrete ratings with values in $\cS:=\{1,\ldots,5\}$ examined by $D=943$ users for $T=1,682$ movies~\cite{MAL03}. To highlight the merits of the novel CSL schemes, a fraction $p$ of the ratings were randomly sampled as training data to learn the latent subspace, and sketching was performed using our scheme and the MM one. Dimension $d=8$ is selected for the latent subspace. Due to the small size of the training dataset, a single pass may lead to unsatisfactory learning accuracy when initialized randomly. Hence, to improve the ability of our scheme to learn the subspace, three passes were allowed over the data, where the first pass was initialized randomly, and then the resulting subspace formed an initial value for the next round, and so on. It appears that three rounds suffice to attain the learning accuracy of the batch counterpart with reduced computational complexity. The resulting subspace and sketch are then used to interpolate the missing ratings. The runtime and root-mean-square-error ${\rm RMSE}=\Big(\frac{1}{T} \sum_{t=1}^T \|\by_t-\hat{\by}_t\|^2\Big)^{1/2}$ are listed in Table \ref{tab:movielens_result}. It is seen that the novel approach outperforms the MM scheme in terms of both runtime and prediction accuracy. For instance, with $30\%$ missing ratings our scheme offers around $5\%$ gain in prediction accuracy with three times lower runtime.

{\color{red}

\begin{table}
\begin{center}
 \begin{tabular}[H]{ c| c | c| c| c }
\hline
    & \multicolumn{2}{c|}{ML-online} & \multicolumn{2}{c}{MM [10]}\\ \hline
   $p$  & runtime &RMSE&runtime&RMSE \\ \hline
    0.7 & 4.2702 &0.8688 &5.4340 &0.8947\\ \hline
    0.8&  4.0327  &  0.8549& 5.0281&0.8944 \\ \hline
    0.9&  4.1744& 0.8441 &5.2766& 0.8936\\
\hline
    \end{tabular}
\end{center}
\caption{Runtime and RMSE comparison of the proposed against the batch MM scheme under various $p$ for MovieLens dataset (D2) with $d=8$.}\label{tab:movielens_result}
\end{table}
}

%

%
%
\begin{figure*}[t]
\begin{minipage}[b]{.52\textwidth}
\centering
\hspace{-2mm}\includegraphics[width=8.8cm]{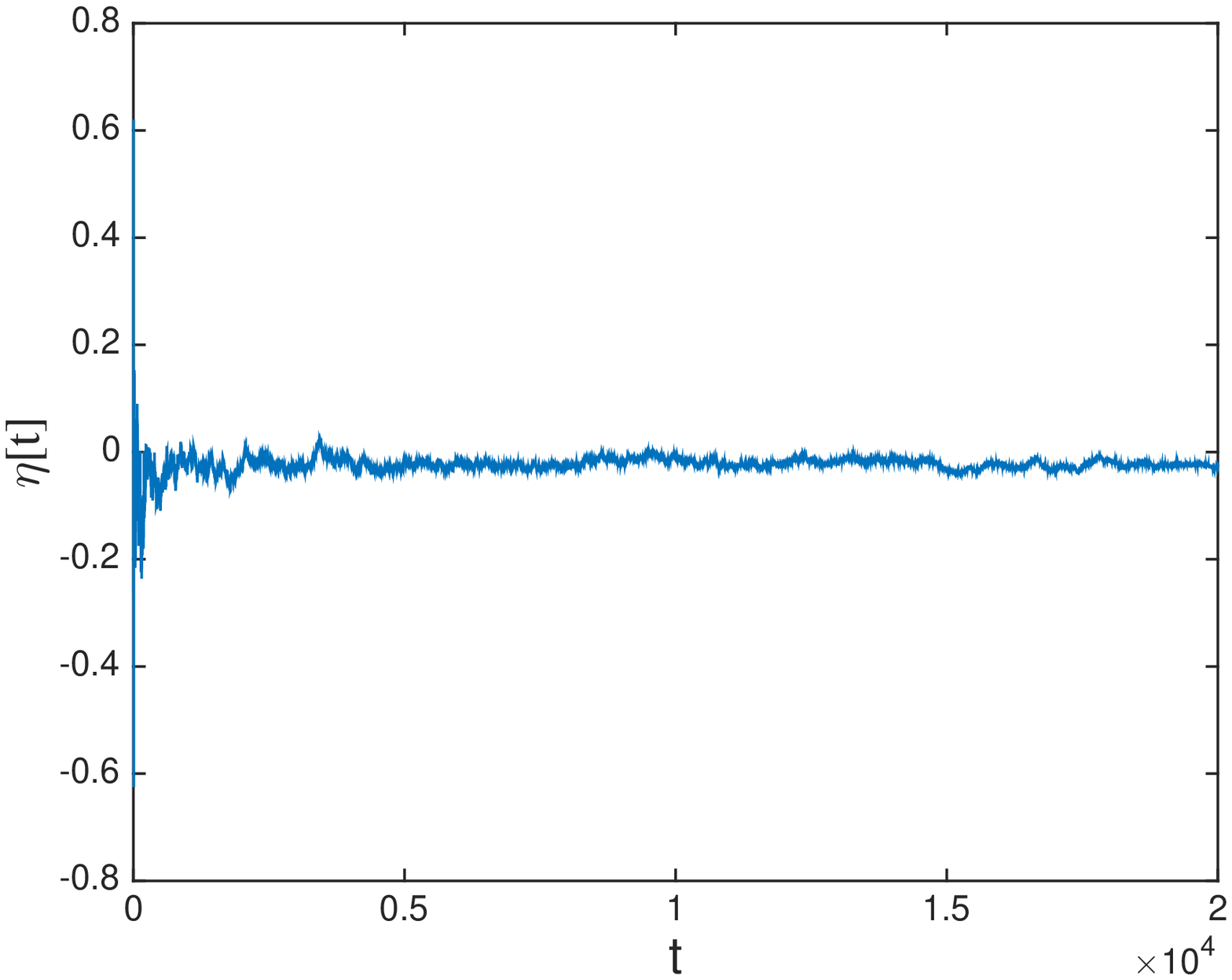}
\hspace{-2mm}
\centerline{(a)}
\end{minipage}
\begin{minipage}[b]{.52\textwidth}
\centering
\hspace{-5mm}\includegraphics[width=8.8cm]{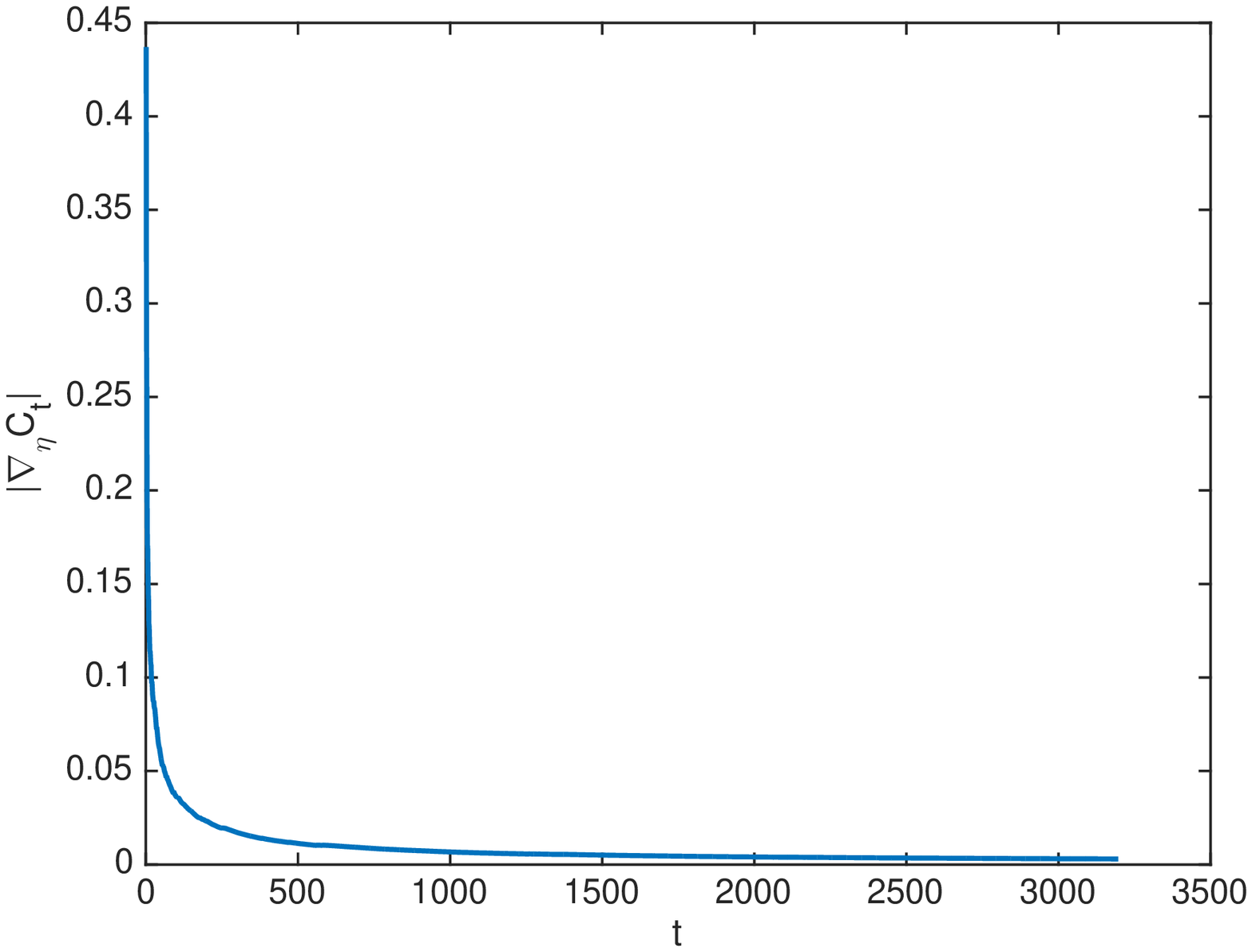}
\hspace{-2mm}
\centerline{(b)}
\end{minipage}
 \caption{Quantization threshold convergence; (left) threshold evolution, and (right) threshold gradient absolute value evolution for chess data. } 
 \label{fig:tau}
\end{figure*}

\subsection{Threshold adaptation}

In this section, convergence and effectiveness of our quantization threshold adaptation is tested for the binary synthetic data described in Sec.~\ref{subsec:syn}. It is observed from Fig.~\ref{fig:tau}(a) that by learning $\eta$, the threshold approaches the ground-truth value of $\eta=0$. The interpolation error as well as the SVM-classification error using the resulting sketch are reported in Table \ref{tab:syn_tau}. Clearly, the threshold adaptation improves the interpolation accuracy by about $17 \%$ relative to the CSL scheme that uses the fixed threshold $\eta=0.5$. 

Threshold adaptation is also evaluated on the real chess-game data classification. The performance reported in Table \ref{tab:chess_tau} shows again $3.7 \%$ accuracy improvement relative to the non-adaptive scheme. It is also empirically observed in Fig.~\ref{fig:tau}(b) that with the joint quantization threshold and CSL, the threshold iterates converge to a stationary point of the nuclear-norm regularized ML estimator.

\begin{table}
\begin{center}
	\vspace{0.2cm}
 \begin{tabular}[t]{ c| c | c| c }
\hline
\hline
    & \multicolumn{3}{c}{online CSL} \\ \hline
   $p$  & Runtime (sec) & RMSE & classification error ($\%$)\\ \hline
   0.6 & 4.4117 & 0.3464 & 6.57\\ \hline
    0.7 & 4.4146 &0.3341 & 6.02\\ \hline
    0.8 &  4.4782  &  0.2910 & 4.64\\ \hline
    0.9 &  5.8252 & 0.2792 & 4.07\\ \hline
    
    & \multicolumn{3}{c}{online CSL with threshold adaptation}\\ \hline
    0.6 & 4.8325 & 0.2967 & 6.32\\ \hline
    0.7& 4.7555 & 0.2846 & 5.19\\ \hline
    0.8& 4.6931 & 0.2737 & 4.52\\ \hline
    0.9& 5.1522 & 0.2668 & 3.69\\ \hline
\hline
    \end{tabular}
\end{center}
	\vspace{1mm}
\caption{RMSE and classification accuracy comparison of the novel CSL scheme with, and without threshold adaptation, under various $p$ for binary synthetic data when $d=5$, $D=20$, and $T=5,000$.}\label{tab:syn_tau}
\end{table}

\begin{table}
\begin{center}
	\vspace{0.2cm}
 \begin{tabular}[t]{ c| c | c| c }
\hline
\hline
    & \multicolumn{3}{c}{online CSL } \\ \hline
   $p$  & Runtime (sec) & RMSE & classification error (\%)\\ \hline
   0.6 & 1.5521 & 0.7751 & 24.62\\ \hline
    0.7 & 1.7344 & 0.7740 & 24.59\\ \hline
    0.8 &  1.7949  &  0.7736 & 24.52\\ \hline
    0.9&  2.1000& 0.7729 & 24.36\\ \hline
    & \multicolumn{3}{c}{online CSL with threshold adaptation}\\ \hline
    0.6 & 1.8037 & 0.7725 & 23.73\\ \hline
    0.7& 2.2913 & 0.7724 & 23.67\\ \hline
    0.8& 2.1271 & 0.7729 & 23.31\\ \hline
    0.9& 2.2210 & 0.7708 & 23.16\\ \hline
\hline
    \end{tabular}
\end{center}
\vspace{1mm}
\caption{RMSE and classification accuracy comparison of the CSL scheme with, and without threshold adaptation, under various $p$ for the chess-game dataset when $d=5$, $D=35$, and $T=3,196$.}\label{tab:chess_tau}
\end{table}

\section{Conclusions and Future Directions}
\label{sec:conc}
Effective sketching approaches were developed in this paper for large-scale categorical data that are incomplete and streaming. Low-dimensional Probit, Tobit and Logit models were considered and learned, using a maximum likelihood approach regularized with a surrogate of the nuclear norm. Leveraging separability of this regularizer, and employing stochastic alternating minimization, online algorithms were subsequently developed to sketch the data `on the fly.' The resultant learning task refines the latent subspace upon arrival of a new datum, and then forms the sketch by projecting the imputed datum onto the latent subspace. This leads to first-order, lightweight, and parallelized iterations. The quantization thresholds are also learned along with the subspace to enhance the modeling flexibility. Performance of the novel algorithms was assessed for both infinite and finite data streams, where for the former asymptotic convergence was established, while for the latter sublinear regret bounds were derived. Simulated tests were carried out on both synthetic and real datasets to confirm the efficacy of the novel schemes for real-time movie recommendation and chess-game classification tasks. 

There are still intriguing questions beyond the scope of the present study, that are worth pursuing as future research. One direction pertains to utilizing kernels for nonlinear subspace modeling in an online and computationally efficient fashion. Improving robustness of the categorical subspace learning for dynamic environments with time-varying subspaces is another important avenue to explore.


\appendix

\noindent\textbf{Proof of Lemma~\ref{lemma1}}:~Assuming $y_{i,t}=\eta_j$ without loss of generality, gradient and Hessian are first derived in closed form
	\begin{align}
	&\nabla_{\bbu_i}g^{\rm (probit)}_t\big(\bbpsi_{t},\mathbf{U}\big)\nonumber\\
	& = -\sigma^{-1} \left[ \frac{ {\rm \phi}(z^i_{j,t-1})-{\rm \phi}(z^i_{j+1,t-1})}{ {\rm Q}(z^i_{j,t-1})-{\rm Q}(z^i_{j+1,t-1})}\right] \bbpsi_t
+\frac{\lambda}{t}\bbu_i\label{eq:h:grad}
	\\
	&\nabla^2_{\bbu_i} g^{\rm (probit)}_t\big(\bbpsi_{t},\mathbf{U}\big) =\sigma^{-2}\bigg\{\left[\frac{{\rm \phi}(z^i_{j,t-1})-{\rm \phi}(z^i_{j+1,t-1})}{{\rm Q}(z^i_{j,t-1})-{\rm Q}(z^i_{j+1,t-1})}\right]^2\nonumber\\
	&-\frac{z^i_{j,t-1}{\rm \phi}(z^i_{j,t-1})-z^i_{j+1,t-1}{\rm \phi}(z^i_{j+1,t-1})}{\left[{\rm Q}(z^i_{j,t-1})-{\rm Q}(z^i_{j+1,t-1})\right]} \bigg\} \bbpsi_t\bbpsi_t^T+\frac{\lambda}{t}\bbI\label{eq:h:hessian}
	\end{align}
	where $z_{j,t-1}^i:={\sigma}^{-1}(\eta_{j}-\mathbf{u}_i^{\top}[t-1]\bbpsi_t)$. Let us also define 
	\begin{align}
	r_j:&=  -\frac{ {\rm \phi}(z^i_{j,t-1})-{\rm \phi}(z^i_{j+1,t-1})}{{\rm Q}(z^i_{j,t-1})-{\rm Q}(z^i_{j+1,t-1})}\nonumber\\
	&=\frac{1}{{\rm Q}(z^i_{j,t-1})-{\rm Q}(z^i_{j+1,t-1})}\int_{z^i_{j,t-1}}^{z^i_{j+1,t-1}}\epsilon{\rm \phi}(\epsilon)d\epsilon.
	\end{align}
	Since~$z^i_{j,t-1}<z^i_{j+1,t-1}$, we have
	\begin{align}
	&z^i_{j,t-1}({\rm Q}\left(z^i_{j,t-1}\right)-{\rm Q}(z^i_{j+1,t-1}))
	\leq\int_{z^i_{j,t-1}}^{z^i_{j+1,t-1}}\epsilon {\rm \phi}(\epsilon)d\epsilon\nonumber\\
	&\leq z^i_{j+1,t-1}({\rm Q}\left(z^i_{j,t-1}\right)-{\rm Q}(z^i_{j+1,t-1}))
	\end{align}
	%
and therefore, 
	\begin{align}
	\label{eq:rj}
	r_j \in [z^i_{j,t-1},z^i_{j+1,t-1}]\leq\sigma^{-1}(\eta_j-\eta_{j-1}).
	\end{align}
	 Hence, one can simply bound the gradient as~$\|\nabla_{\bbu_i} g_t\big(\bbpsi_{t},\mathbf{U}\big)\|_2\leq \left\|{(\eta_{J-1}-\eta_1)}\bbpsi_t/{\sigma^2} +{\lambda}\bbu_i/t\right\|_2$. Resorting to the triangle inequality, we obtain
	 \begin{align}
	 	\|\nabla_{\bbu_i} g^{\rm (probit)}_t\big(\bbpsi_{t},\mathbf{U}\big)\|_2\leq \delta_1 \|\bbpsi_t\|_2+\frac{\lambda}{t}\|\bbu_i\|_2
	 \end{align}
	  where $\delta_1:=\Delta/\sigma^2$, and $\Delta:=\eta_{J-1}-\eta_{0}$ is the quantization range.

Likewise, we have
	\begin{align}
	&\hspace{-8mm} -\frac{z^i_{j,t-1} {\rm \phi}(z^i_{j,t-1})-z^i_{j+1,t-1} {\rm \phi}(z^i_{j+1,t-1})}{\left[{\rm Q}(z^i_{j,t-1})-{\rm Q}(z^i_{j+1,t-1})\right]}\nonumber\\
	=&1-\frac{1}{{\rm Q}(z^i_{j,t-1})-{\rm Q}(z^i_{j+1,t-1})}\int_{z^i_{j,t-1}}^{z^i_{j+1,t-1}}\epsilon^2 {\rm \phi}(\epsilon)d\epsilon\leq 1\nonumber
	\end{align}
	which implies that the Hessian can simply be bounded by
	\begin{align}
	\|\nabla_{\bbu_i}^2g_t^{\rm (probit)}\big(\bbpsi_{t},\mathbf{U}\big)\|\leq \frac{r_j^2+1}{\sigma^2}\|\bbpsi_t\|^2_2+\frac{\lambda}{t}	\end{align}
and thus, 
\begin{align}
	\|\nabla_{\bbu_i}^2 h^{\rm (probit)}_t\big(\bbpsi_{t},\mathbf{U}\big)\|\leq \delta_2\|\bbpsi_t\|_2^2+\frac{\lambda }{t}
\end{align}
where $\delta_2:=(\Delta^2/\sigma^2+1)/\sigma^2$.
Hence, the compactness assumption (as2) implies that the gradient and Hessian are bounded. The differentiability of  $g_t$ then leads to Lipschitz continuity of $g_t$ and $\nabla g_t$.

	
		%
	%

\noindent\textbf{Proof of Lemma \ref{lemma2}:}~According to the gradient expression in \eqref{eq:gu:probit}, the Hessian for the Probit cost function can be written as
\begin{multline}
\nabla_{\bbu_i}^{2}g_t^{\rm (probit)}\big(\bbpsi_{t},\mathbf{U}\big)=\bigg\{\left[\frac{f(\bbu_i,\bbpsi_t)}{w(\bbu_i,\bbpsi_t)} \right]^2 \\
-\frac{m(\bbu_i,\bbpsi_t)}{w(\bbu_i,\bbpsi_t)}\bigg\} \bbpsi_t\bbpsi_t^{\top}+\frac{\lambda}{t}\bbI\label{eq:hessian}
\end{multline}
where 
\begin{eqnarray}
m(\bbu_i,\bbpsi_t)&:=&z^i_{j,t-1} {\rm \phi}(z^i_{j,t-1})-z^i_{j+1,t-1} {\rm \phi}(z^i_{j+1,t-1})\nonumber\\
 f(\bbu_i,\bbpsi_t)&:=&{\rm \phi}(z^i_{j,t-1})-{\rm \phi}(z^i_{j+1,t-1})\nonumber\\
 w(\bbu_i,\bbpsi_t)&:=&{\rm Q}\left(z^i_{j,t-1}\right)-{\rm Q}(z^i_{j+1,t-1})\;\nonumber
\end{eqnarray}
From \eqref{eq:rj} and the definition of $m(\bbu_i[t-1],\bbpsi_t)$, we have
\begin{align}
\label{ineq:m}
{z^i_{j,t-1} f(\bbu_i,\bbpsi_t)}\leq m(\bbu_i,\bbpsi_t)\leq {z^i_{j+1,t-1} f(\bbu_i,\bbpsi_t)}.
\end{align}
If  $r_j>0$, then $z^i_{j+1,t-1}>0$, which in combination with \eqref{ineq:m} yields
\begin{multline}
\left[\frac{f(\bbu_i,\bbpsi_t)}{w(\bbu_i,\bbpsi_t)}\right]^2
-\frac{m(\bbu_i,\bbpsi_t)}{w(\bbu_i,\bbpsi_t)} \\
\geq  r_j^2-z^i_{j,t-1}r_j=r_j(r_j-z^i_{j,t-1})\geq 0.\label{ineq:cvx1}
\end{multline}
Similarly, if $r_j<0$, it follows that
\begin{multline}
\left[\frac{f(\bbu_i,\bbpsi_t)}{w(\bbu_i,\bbpsi_t)} \right]^2
-\frac{m(\bbu_i,\bbpsi_t)}{w(\bbu_i,\bbpsi_t)}\\
\geq r_j^2-z^i_{j+1,t-1}r_j=r_j(r_j-z^i_{j+1,t-1})\geq 0. \label{ineq:cvx2}
\end{multline}
Clearly \eqref{ineq:cvx1} and \eqref{ineq:cvx2} imply that the Hessian matrix in \eqref{eq:hessian} is positive definite. Hence, the entry-wise cost $g_{t}(\cdot) $  is convex w.r.t. $\bbu_i$. Likewise, due to its symmetry w.r.t. $\bu_i$ and $\bpsi_t$, the cost $g_t(\cdot)$ is convex w.r.t. $\bpsi_t$.


For the binary Logit model, the Hessian of the function can be represented as (cf. \eqref{eq:gu:probit})
\begin{eqnarray}
	&&\hspace{-17mm}\nabla_{\bbu_i}^{2}g_t^{\rm (logit)}\big(\bbpsi_{t},\mathbf{U}\big)\nonumber\\
	&=&\bigg\{\frac{(2y_{i,t}-1)^2\exp(\bbu_i^\top\bbpsi_t)}{1+\exp((2y_{i,t}-1)\bbu_i^\top\bbpsi_t)}\bigg\} \bbpsi_t\bbpsi_t^\top+\frac{\lambda}{t}\bbI\nonumber\\
	&=&\bigg\{\frac{\exp(\bbu_i^\top\bbpsi_t)}{1+\exp((2y_{i,t}-1)\bbu_i^\top\bbpsi_t)}\bigg\}\bbpsi_t\bbpsi_t^\top+\frac{\lambda}{t}\bbI\label{eq:hessian:logit}
\end{eqnarray}
where the last equation comes from the fact that $|2y_{i,t}-1|=1$. It is clear that
\begin{align}
\frac{\exp(\bbu_i^\top\bbpsi_t)}{1+\exp((2y_{i,t}-1)\bbu_i^{\top}\bbpsi_t)}>0
\end{align}
and hence $\nabla_{\bbu_i}^{2}g_t^{(\rm logit)}\big(\bbpsi_{t},\mathbf{U}\big)\succ \mathbf{0}$. Likewise, the Hessian matrix of $\bbpsi$ for a fixed subspace $\bU$ is also positive definite because the objective function is symmetric with respect to $\bbu_i$ and $\bbpsi_t$. Hence, the entry-wise cost function is per-block convex in terms of $\bbu_i$ and $\bbpsi_t$.

For the Tobit-II model in \eqref{eq:gu:tobit2}, the gradient looks similar to that of the Probit model for $y_{i,t}\in(\eta_l,\eta_u)$, and the only difference appears in the threshold values, which will not influence convexity of the function. In fact, for $y_{i,t}=\eta_u$ or $y_{i,t}=\eta_l$, we arrive at
\begin{align}
	\nabla_{\bbu_i}^{2}g_t^{\rm (tobit-II)}\big(\bbpsi_{t},\mathbf{U}\big) =\frac{1}{\sigma^2}\bbpsi_t\bbpsi_t^{\top}+\frac{\lambda}{t}\bbI
\end{align}
which is positive definite. Likewise, the Hessian matrix of $\bbpsi$ for a fixed $\bU$ is also positive definite due to the symmetry of $\bbu_i$ and $\bbpsi_t$. Hence, the entry-wise cost is per-block convex in terms of $\bbu_i$ and $\bbpsi_t$.

\noindent\textbf{Proof of Lemma \ref{lem:lemma3}:}
First, observe that $\nabla\bar{C}_t(\bbU[t]) =\nabla\bar{C}_{t+1}(\bbU[t+1])=\boldsymbol{{0}}$ by construction of the algorithm. Meanwhile, since $\bar{C}_t(\bbU)$ is strongly convex (cf. Lemma \ref{lemma2}), the mean-value theorem implies 
	\begin{align}
		\bar{C}_t(\bbU[t+1])&\geq \bar{C}_t(\bbU[t])+\frac{\rho}{2}\big\|\bbU[t+1]-\bbU[t]\big\|_F^2\nonumber\\
		\bar{C}_{t+1}(\bbU[t])&\geq \bar{C}_{t+1}(\bbU[t+1])+\frac{\rho}{2}\big\|\bbU[t+1]-\bbU[t]\big\|_F^2\nonumber
	\end{align}
	where $\rho$ denotes the strong convexity constant of $\bar{C}_t(\bbU[t+1])$. Upon defining the function $\nu_t(\bbU):=\bar{C}_t(\bbU)-\bar{C}_{t+1}(\bbU)$, we arrive at
	\begin{align}
	\big\|\bbU[t+1]-\bbU[t]\big\|_F^2
	\leq  \frac{1}{\rho}\big|\nu_t(\bbU[t+1])-\nu_t(\bbU[t])\big|.\label{ineq:2}
	\end{align}
	Based on the definition of $\bar{C}(\bbU[t+1])$, we further have
	{\small
	\begin{align}
	\nu_t(\bbU)&=\frac{1}{t}\sum_{\tau=1}^t g_{\tau}(\bbpsi_{\tau},\bbU)-\frac{1}{t+1}\sum_{\tau=1}^{t+1} g_{\tau}(\bbpsi_{\tau},\bbU)\nonumber\\
	&= \frac{1}{t(t+1)}\sum_{\tau=1}^{t}g_{\tau}(\bbpsi_{\tau},\bbU)-\frac{1}{t+1}g_{t+1}(\bbpsi_{\tau+1},\bbU).\label{def:u}
	\end{align}
	}
	Combining Lemma \ref{lemma1} with \eqref{def:u}, establishes that $\nu_t(\bbU)$ is Lipschitz continuous, and thus
	
	{\small
    \begin{align}
	\big|\nu_t(\bbU[t+1])-\nu_t(\bbU[t])\big| \leq \frac{ \lambda B_u+\delta_1 B_{\psi}}{t+1}\big\|\bbU[t+1]-\bbU[t]\big\|_F
	\end{align}
	}
	which after using \eqref{ineq:2} yields
	\begin{align}
		\big\|\bbU[t+1]-\bbU[t]\big\|_F\leq \frac{\lambda B_u+\delta_1 B_{\psi}}{(t+1)\rho}.
	\end{align}
	Accordingly, Lemma \ref{lem:lemma3} holds with $B:={ (\lambda B_u+\delta_1 B_{\psi})}/{\rho}$.

\bibliography{mybib}
\bibliographystyle{IEEEtranS}
\end{document}